\documentclass[runningheads]{llncs}
\usepackage{graphicx}
\usepackage{subfig}
\usepackage{amsmath,amssymb} 
\usepackage{color}
\usepackage[width=122mm,left=12mm,paperwidth=146mm,height=193mm,top=12mm,paperheight=217mm]{geometry}
\usepackage{multirow}
\newcommand{\tabincell}[2]{\begin{tabular}{@{}#1@{}}#2\end{tabular}}
\usepackage[breaklinks=true,letterpaper=true,colorlinks,bookmarks=false]{hyperref}

\begin{document}

\pagestyle{headings}
\mainmatter

\title{ESRGAN: Enhanced Super-Resolution Generative Adversarial Networks}

\titlerunning{ESRGAN: Enhanced Super-Resolution Generative Adversarial Networks}

\authorrunning{Xintao Wang \textit{et al.}}

\author{
	Xintao Wang$^1$, Ke Yu$^1$, Shixiang Wu$^2$, Jinjin Gu$^3$, Yihao Liu$^4$, \\
	Chao Dong$^2$, Chen Change Loy$^5$, Yu Qiao$^2$, Xiaoou Tang$^1$
}
\institute{
	\scriptsize
	$^1$CUHK-SenseTime Joint Lab, The Chinese University of Hong Kong \\
	$^2$SIAT-SenseTime Joint Lab, Shenzhen Institutes of Advanced Technology,\\Chinese Academy of Sciences
	$^3$The Chinese University of Hong Kong, Shenzhen \\
	$^4$University of Chinese Academy of Sciences
	$^5$Nanyang Technological University, Singapore \\
	\email{ \{wx016,yk017,xtang\}@ie.cuhk.edu.hk, \{sx.wu,chao.dong,yu.qiao\}@siat.ac.cn 
		liuyihao14@mails.ucas.ac.cn, 115010148@link.cuhk.edu.cn, ccloy@ntu.edu.sg }
}

\maketitle

\begin{abstract}
	The Super-Resolution Generative Adversarial Network (SRGAN)~\cite{ledig2017photo} is a seminal work that is capable 
	of generating realistic textures during single image super-resolution.
	However, the hallucinated details are often accompanied with unpleasant artifacts.
	To further enhance the visual quality, we thoroughly study three key components of SRGAN -- network architecture, 
	adversarial loss and perceptual loss, and improve each of them to derive an Enhanced SRGAN (ESRGAN).
	In particular, we introduce the Residual-in-Residual Dense Block (RRDB) \mbox{without} batch normalization as the 
	basic network building unit.
	Moreover, we borrow the idea from relativistic GAN~\cite{jolicoeur2018relativistic} to let the discriminator 
	predict relative realness instead of the absolute value.
	Finally, we improve the perceptual loss by using the features before activation, which could provide stronger 
	supervision for brightness consistency and texture recovery. 
	Benefiting from these improvements, the proposed ESRGAN achieves consistently better visual quality with more 
	realistic and natural textures than SRGAN and won the first place in the PIRM2018-SR Challenge\footnote{We 
	won the first place in region 3 and got the best perceptual index.}~\cite{pirm18url}. 
	The code is available at \url{https://github.com/xinntao/ESRGAN}.
	
\end{abstract}

\section{Introduction}

\begin{figure}[htbp]
	\begin{center}
		\includegraphics[width=\linewidth]{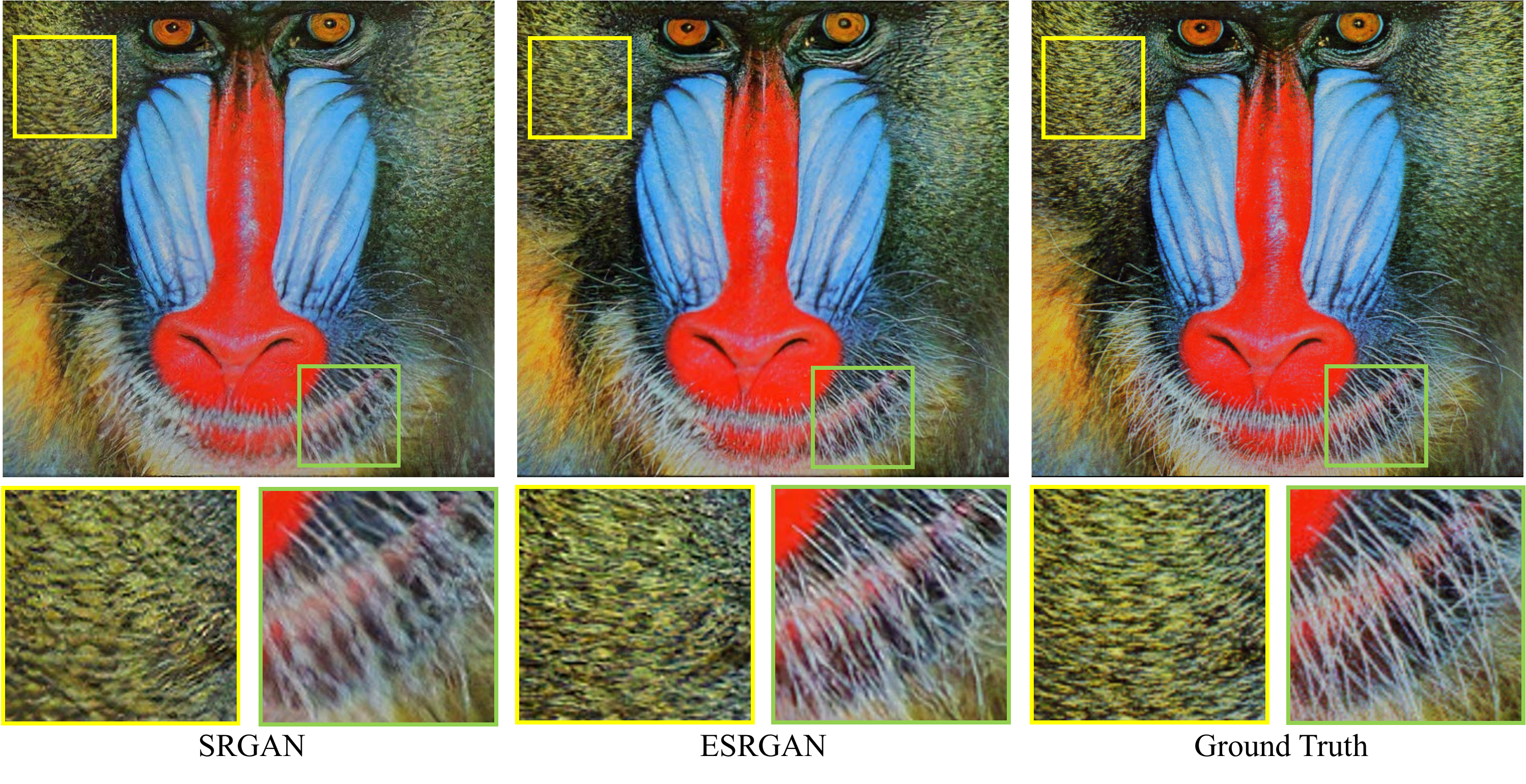}
	\end{center}
	\vspace{-0.4cm}
	\caption{The super-resolution results of $\times 4$ for SRGAN\protect\footnotemark, the proposed ESRGAN and the 
		ground-truth. ESRGAN outperforms SRGAN in sharpness and details.}
	\vspace{-0.4cm}
	\label{fig:tease}
\end{figure}

Single image super-resolution (SISR), as a fundamental low-level vision problem, has attracted increasing attention in 
the research community and AI companies.
SISR aims at recovering a high-resolution (HR) image from a single low-resolution (LR) one.
Since the pioneer work of SRCNN proposed by Dong et al.~\cite{dong2014learning}, deep convolution neural network 
(CNN) approaches have brought prosperous development.
Various network architecture designs and training strategies have continuously improved the SR performance, especially 
the Peak Signal-to-Noise Ratio (PSNR) 
value~\cite{kim2016accurate,lai2017deep,kim2016deeply,ledig2017photo,tai2017image,tai2017memnet,haris2018deep,zhang2018residual,zhang2018image}.
However, these PSNR-oriented approaches tend to output over-smoothed results without sufficient high-frequency details, 
since the PSNR metric fundamentally disagrees with the subjective evaluation of human observers~\cite{ledig2017photo}.

Several perceptual-driven methods have been proposed to improve the visual quality of SR results.
For instance, perceptual loss~\cite{johnson2016perceptual,bruna2015super} is proposed to optimize super-resolution 
model in a feature space instead of pixel space. 
Generative adversarial network~\cite{goodfellow2014generative} is introduced to SR 
by~\cite{ledig2017photo,sajjadi2017enhancenet} to encourage the network to favor solutions that look more like natural 
images.
The semantic image prior is further incorporated to improve recovered texture details~\cite{wang2018sftgan}.
One of the milestones in the way pursuing visually pleasing results is SRGAN~\cite{ledig2017photo}.
The basic model is built with residual blocks~\cite{he2016deep} and optimized using perceptual loss in a GAN framework.
With all these techniques, SRGAN significantly improves the overall visual quality of reconstruction over PSNR-oriented 
methods.

\footnotetext{We use the released results of original SRGAN~\cite{ledig2017photo} paper -- 
	\url{https://twitter.app.box.com/s/lcue6vlrd01ljkdtdkhmfvk7vtjhetog}.}

However, there still exists a clear gap between SRGAN results and the ground-truth (GT) images, as shown in 
Fig.~\ref{fig:tease}. 
In this study, we revisit the key components of SRGAN and improve the model in three aspects.
First, we improve the network structure by introducing the Residual-in-Residual Dense Block (RDDB), which is of higher 
capacity and easier to train.
%
We also remove Batch Normalization (BN)~\cite{ioffe2015batch} layers as in~\cite{lim2017enhanced} and use residual 
scaling~\cite{szegedy2016inception,lim2017enhanced} and smaller initialization to facilitate training a very deep 
network. 
Second, we improve the discriminator using Relativistic average GAN (RaGAN)~\cite{jolicoeur2018relativistic}, which 
learns to judge ``whether one image is more realistic than the other" rather than ``whether one image is real or 
fake". 
Our experiments show that this improvement helps the generator recover more realistic texture details.
Third, we propose an improved perceptual loss by using the VGG features \textit{before activation} instead of after 
activation as in SRGAN.
We empirically find that the adjusted perceptual loss provides sharper edges and more visually pleasing results, as 
will be shown in Sec.~\ref{subsec:ablation_study}.
Extensive experiments show that the enhanced SRGAN, termed ESRGAN, consistently outperforms state-of-the-art methods in 
both sharpness and details (see Fig.~\ref{fig:tease} and Fig.~\ref{fig:qualitative_results}). 

We take a variant of ESRGAN to participate in the PIRM-SR Challenge~\cite{pirm18url}.
This challenge is the first SR competition that evaluates the performance in a perceptual-quality aware manner based 
on~\cite{blau2017perception}, where the authors claim that distortion and perceptual quality are at odds with each 
other. 
The perceptual quality is judged by the non-reference measures of Ma's score~\cite{ma2017learning} and 
NIQE~\cite{mittal2013making}, i.e., perceptual index = $\frac{1}{2} ((10-\text{Ma})+\text{NIQE})$. 
A lower perceptual index represents a better perceptual quality.

\begin{figure}[tbp]
	\begin{center}
		\includegraphics[width=0.9\linewidth]{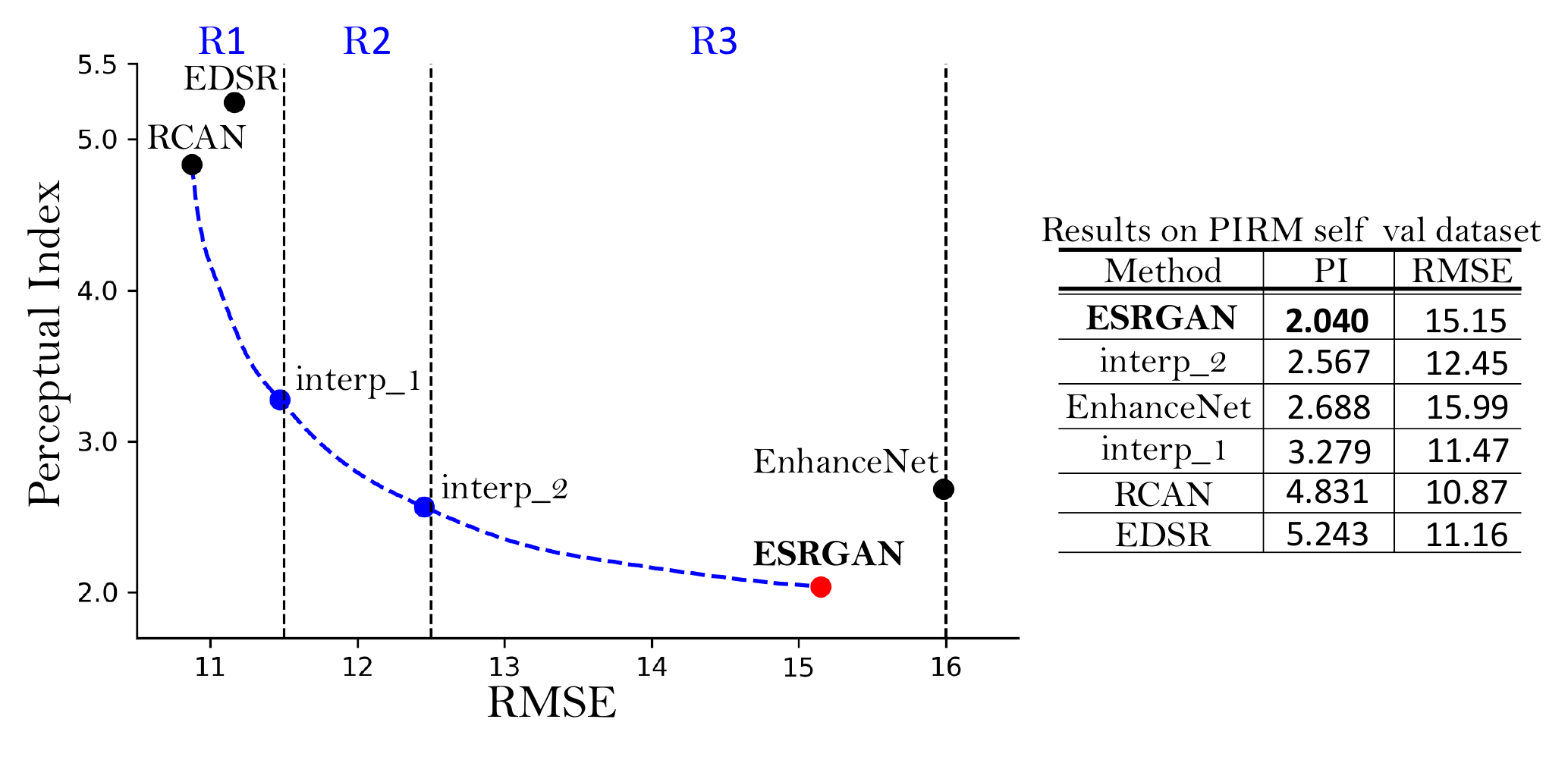}
	\end{center}
	\vspace{-0.4cm}
	\caption{Perception-distortion plane on PIRM self validation dataset. We show the baselines of 
		EDSR~\cite{lim2017enhanced}, RCAN~\cite{zhang2018image} and EnhanceNet~\cite{sajjadi2017enhancenet}, and the 
		submitted ESRGAN model. The blue dots are produced by image interpolation.}
	\label{fig:perception_distortion_plane}
	\vspace{-0.4cm}
\end{figure}

As shown in Fig.~\ref{fig:perception_distortion_plane}, the perception-distortion plane is divided into three regions 
defined by thresholds on the Root-Mean-Square Error (RMSE), and the algorithm that achieves the lowest perceptual index 
in each region becomes the regional champion.
We mainly focus on region 3 as we aim to bring the perceptual quality to a new high. 
Thanks to the aforementioned improvements and some other adjustments as discussed in 
Sec.~\ref{sec:prim18}, our proposed ESRGAN won the first place in the PIRM-SR Challenge (region 3) with the best 
perceptual index.

In order to balance the visual quality and RMSE/PSNR, we further propose the network interpolation strategy, which 
could continuously adjust the reconstruction style and smoothness.
%
Another alternative is image interpolation, which directly interpolates images pixel by pixel. 
We employ this strategy to participate in region 1 and region 2. 
The network interpolation and image interpolation strategies and their differences are discussed in 
Sec.~\ref{subsec:net_interp}.

\section{Related Work}
We focus on deep neural network approaches to solve the SR problem.
As a pioneer work, Dong et al.~\cite{dong2014learning,dong2016image} propose SRCNN to learn the mapping from LR to HR 
images in an end-to-end manner, achieving superior performance against previous works.
Later on, the field has witnessed a variety of network architectures, such as a deeper network with residual
learning~\cite{kim2016accurate}, Laplacian pyramid structure~\cite{lai2017deep}, residual blocks~\cite{ledig2017photo},
recursive learning~\cite{kim2016deeply,tai2017image}, densely connected network~\cite{tai2017memnet}, deep back
projection~\cite{haris2018deep} and residual dense network~\cite{zhang2018residual}.
Specifically, Lim et al.~\cite{lim2017enhanced} propose EDSR model by removing unnecessary BN layers in the residual 
block and expanding the model size, which achieves significant improvement.
Zhang et al.~\cite{zhang2018residual} propose to use effective residual dense block in SR, and they further explore a 
deeper network with channel attention~\cite{zhang2018image}, achieving the state-of-the-art PSNR performance.
Besides supervised learning, other methods like reinforcement learning~\cite{yu2018crafting} and unsupervised 
learning~\cite{yuan2018unsupervised} are also introduced to solve general image restoration problems.

Several methods have been proposed to stabilize training a very deep model. 
For instance, residual path is developed to stabilize the training and improve the 
performance~\cite{he2016deep,kim2016accurate,zhang2018image}.
Residual scaling is first employed by Szegedy et al.~\cite{szegedy2016inception} and also used in EDSR.
For general deep networks, He et al.~\cite{he2015delving} propose a robust initialization method for VGG-style 
networks without BN.
To facilitate training a deeper network, we develop a compact and effective residual-in-residual dense block, 
which also helps to improve the perceptual quality.

Perceptual-driven approaches have also been proposed to improve the visual quality of SR results. 
Based on the idea of being closer to perceptual similarity~\cite{gatys2015texture,bruna2015super}, perceptual 
loss~\cite{johnson2016perceptual} is proposed to enhance the visual quality by minimizing the error in a feature space 
instead of pixel space.
Contextual loss~\cite{roey2018maintaining} is developed to generate images with natural image statistics by using an  
objective that focuses on the feature distribution rather than merely comparing the appearance.
Ledig et al.~\cite{ledig2017photo} propose SRGAN model that uses perceptual loss and adversarial loss to favor 
outputs residing on the manifold of natural images. 
Sajjadi et al.~\cite{sajjadi2017enhancenet} develop a similar approach and further explored the local texture matching 
loss.
Based on these works, Wang et al.~\cite{wang2018sftgan} propose spatial feature transform to effectively incorporate  
semantic prior in an image and improve the recovered textures. 

Throughout the literature, photo-realism is usually attained by adversarial training with  
GAN~\cite{goodfellow2014generative}.
Recently there are a bunch of works that focus on developing more effective GAN frameworks. 
WGAN~\cite{arjovsky2017wasserstein} proposes to minimize a reasonable and efficient approximation of Wasserstein 
distance and regularizes discriminator by weight clipping.
Other improved regularization for discriminator includes gradient clipping~\cite{gulrajani2017improved} and spectral 
normalization~\cite{miyato2018spectral}.
Relativistic discriminator~\cite{jolicoeur2018relativistic} is developed not only to increase the probability that  
generated data are real, but also to simultaneously decrease the probability that real data are real. 
In this work, we enhance SRGAN by employing a more effective relativistic average GAN.

SR algorithms are typically evaluated by several widely used distortion measures, e.g., PSNR and SSIM.
However, these metrics fundamentally disagree with the subjective evaluation of human observers~\cite{ledig2017photo}.
Non-reference measures are used for perceptual quality evaluation, including Ma's score~\cite{ma2017learning} 
and NIQE~\cite{mittal2013making}, both of which are used to calculate the perceptual index in the PIRM-SR 
Challenge~\cite{pirm18url}.
In a recent study, Blau et al.~\cite{blau2017perception} find that the distortion and perceptual quality are at odds 
with each other.

\section{Proposed Methods}

Our main aim is to improve the overall perceptual quality for SR. 
In this section, we first describe our proposed network architecture and then discuss the improvements from the 
discriminator and perceptual loss.
At last, we describe the network interpolation strategy for balancing perceptual quality and PSNR.

\begin{figure}[htbp]
	\vspace{-0.3cm}
	\begin{center}
		\includegraphics[width=\linewidth]{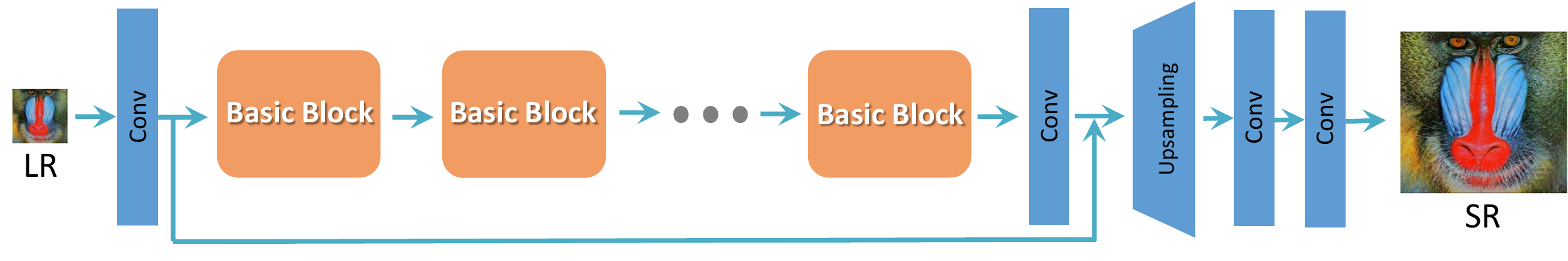}
	\end{center}
	\vspace{-0.4cm}
	\caption{We employ the basic architecture of SRResNet~\cite{ledig2017photo}, where most computation is done in the 
		LR feature space. We could select or design ``basic blocks" (e.g., residual block~\cite{he2016deep}, dense
		block~\cite{huang2016densely}, RRDB) for better performance.}
	\label{fig:generator_architecture}
	\vspace{-0.8cm}
\end{figure}

\subsection{Network Architecture}
\begin{figure}[htbp]
	\vspace{-0.2cm}
	\begin{center}
		\includegraphics[width=\linewidth]{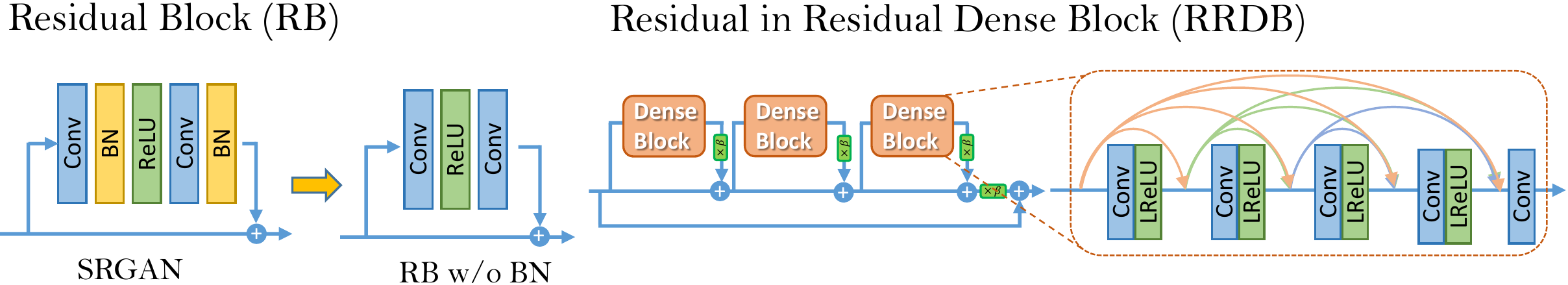}
	\end{center}
	\vspace{-0.3cm}
	\caption{\textbf{Left}: We remove the BN layers in residual block in SRGAN. \textbf{Right}: RRDB block is used in 
		our	deeper model and $\beta$ is the residual scaling parameter.}
	\label{fig:blocks}
	\vspace{-0.3cm}
\end{figure}

In order to further improve the recovered image quality of SRGAN, we mainly make two modifications to the structure 
of generator $G$: 
1) remove all BN layers; 
2) replace the original basic block with the proposed Residual-in-Residual Dense Block (RRDB), which combines 
multi-level residual network and dense connections as depicted in Fig.~\ref{fig:blocks}.

Removing BN layers has proven to increase performance and reduce computational complexity in different PSNR-oriented  
tasks including SR~\cite{lim2017enhanced} and deblurring~\cite{nah2017deep}.
BN layers normalize the features using mean and variance in a batch during training and use estimated mean and 
variance of the whole training dataset during testing.
When the statistics of training and testing datasets differ a lot, BN layers tend to introduce unpleasant artifacts and 
limit the generalization ability. 
We empirically observe that BN layers are more likely to bring artifacts when the network is deeper and trained under 
a GAN framework.
These artifacts occasionally appear among iterations and different settings, violating the needs for a stable 
performance over training.
We therefore remove BN layers for stable training and consistent performance.  
Furthermore, removing BN layers helps to improve generalization ability and to reduce computational complexity and 
memory usage.

%


We keep the high-level architecture design of SRGAN (see Fig.~\ref{fig:generator_architecture}), and use a  
novel basic block namely RRDB as depicted in Fig.~\ref{fig:blocks}.
Based on the observation that more layers and connections could always boost 
performance~\cite{lim2017enhanced,zhang2018residual,zhang2018image}, the proposed RRDB employs a deeper and more 
complex 
structure than the original residual block in SRGAN.
Specifically, as shown in Fig.~\ref{fig:blocks}, the proposed RRDB has a residual-in-residual structure, where residual 
learning is used in different levels.
A similar network structure is proposed in~\cite{zhang2017residual} that also applies a multi-level residual network.
However, our RRDB differs from~\cite{zhang2017residual} in that we use dense block~\cite{huang2016densely}  
in the main path as~\cite{zhang2018residual}, where the network capacity becomes higher benefiting from the dense 
connections.

In addition to the improved architecture, we also exploit several techniques to facilitate training a very deep 
network: 1)  residual scaling~\cite{szegedy2016inception,lim2017enhanced}, i.e., scaling down the residuals by 
multiplying a constant between 0 and 1 before adding them to the main path to prevent instability; 
2) smaller initialization, as we empirically find residual architecture is easier to train when the initial parameter 
variance becomes smaller. More discussion can be found in the \textit{supplementary material}.

The training details and the effectiveness of the proposed network will be presented in Sec.~\ref{sec:exp}.

\subsection{Relativistic Discriminator}

\begin{figure}[htbp]
	\begin{center}
		\includegraphics[width=\linewidth]{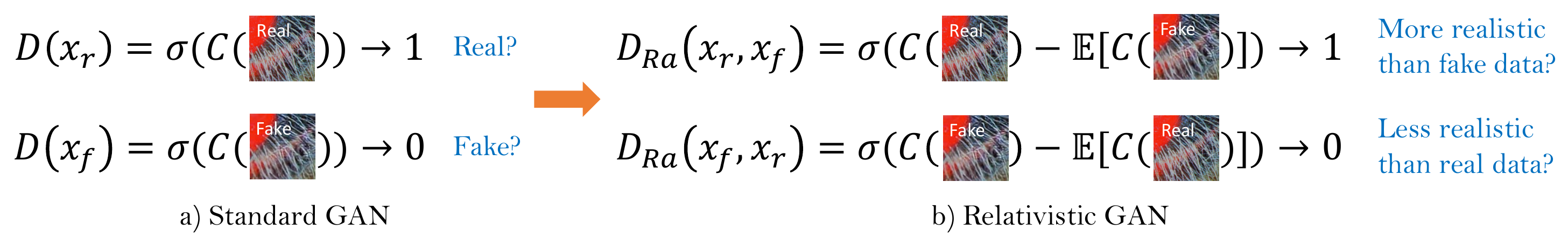}
	\end{center}
	\vspace{-0.4cm}
	\caption{Difference between standard discriminator and relativistic discriminator.}
	\label{fig:RelativisticGAN}
	\vspace{-0.4cm}
\end{figure}

Besides the improved structure of generator, we also enhance the discriminator based on the Relativistic 
GAN~\cite{jolicoeur2018relativistic}. 
Different from the standard discriminator $D$ in SRGAN, which estimates the probability that one input image $x$ is 
real and natural, a relativistic discriminator tries to predict the probability that a real image $x_r$ is relatively 
more realistic than a fake one $x_f$, as shown in Fig.~\ref{fig:RelativisticGAN}.

Specifically, we replace the standard discriminator with the Relativistic average Discriminator 
RaD~\cite{jolicoeur2018relativistic}, denoted as $D_{Ra}$. The standard discriminator in SRGAN can be expressed as 
$D(x)=\sigma(C(x))$, where $\sigma$ is the sigmoid 
function and $C(x)$ is the non-transformed discriminator output. Then the RaD is formulated as $D_{Ra}(x_r, 
x_f)=\sigma(C(x_r)-\mathbb{E}_{x_f}[C(x_f)])$, where $\mathbb{E}_{x_f}[\cdot]$ represents the operation of taking 
average for all fake data in the mini-batch. The discriminator loss is then defined as:
\begin{equation}
L_{D}^{Ra} = -\mathbb{E}_{x_r}[\log(D_{Ra}(x_r, x_f))] - \mathbb{E}_{x_f}[\log(1 - D_{Ra}(x_f, x_r))].
\end{equation}
The adversarial loss for generator is in a symmetrical form:
\begin{equation}
L_{G}^{Ra} = -\mathbb{E}_{x_r}[\log(1 - D_{Ra}(x_r, x_f))] - \mathbb{E}_{x_f}[\log(D_{Ra}(x_f, x_r))],
\end{equation}
where $x_f=G(x_i)$ and $x_i$ stands for the input LR image. 
It is observed that the adversarial loss for generator contains both $x_r$ and $x_f$.
Therefore, our generator benefits from the gradients from both generated data and real data in adversarial training, 
while in SRGAN only generated part takes effect. 
In Sec.~\ref{subsec:ablation_study}, we will show that this modification of discriminator helps to learn sharper edges 
and more detailed textures.

\subsection{Perceptual Loss} \label{subsec:perceptual_loss}

We also develop a more effective perceptual loss $L_\text{percep}$ by constraining on features before activation rather 
than after activation as practiced in SRGAN.

\begin{figure}[htbp]
	\vspace{-0.2cm}
	\begin{center}
		\includegraphics[width=\linewidth]{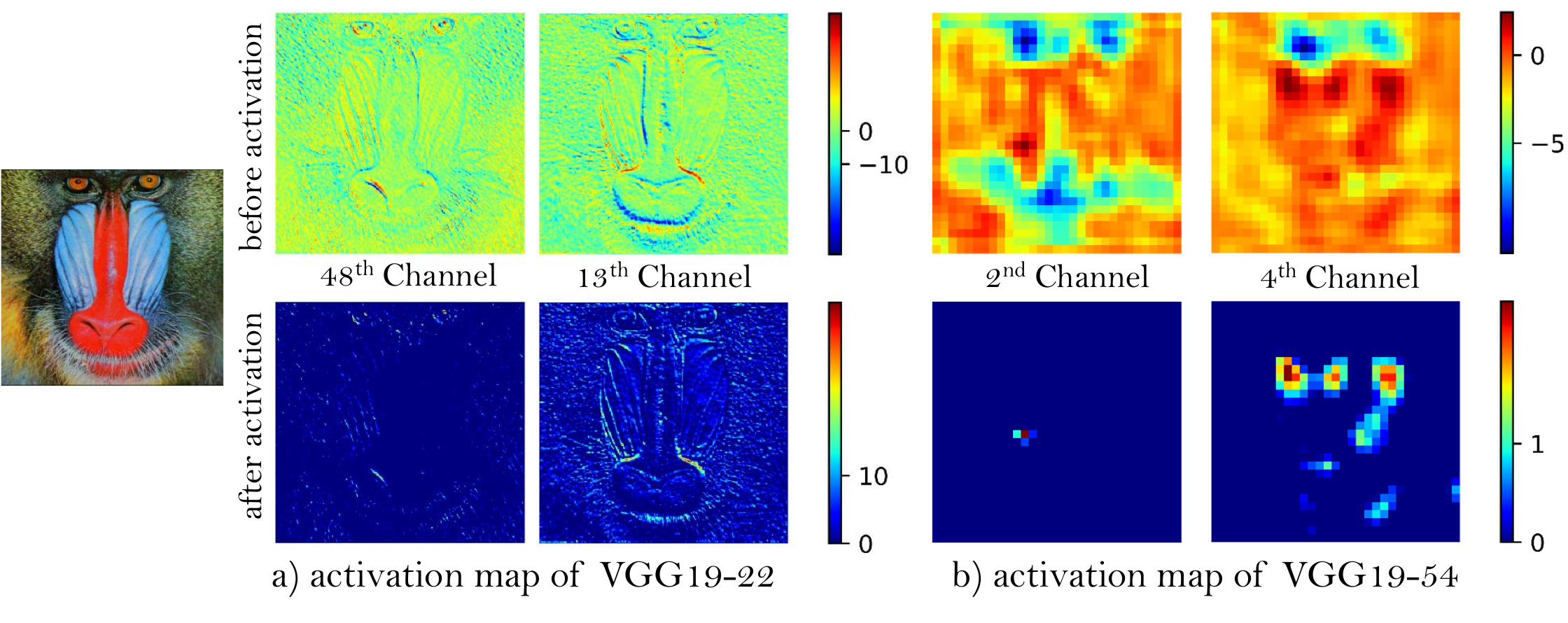}
	\end{center}
	\vspace{-0.5cm}
	\caption{Representative feature maps before and after activation for image `baboon'. With the network going deeper, 
		most of the features after activation become inactive while features before activation contains more 
		information.}
	\label{fig:vgg_sparse}
	\vspace{-0.5cm}
\end{figure}

Based on the idea of being closer to perceptual similarity~\cite{gatys2015texture,bruna2015super}, Johnson et  
al.~\cite{johnson2016perceptual} propose perceptual loss and it is extended in SRGAN~\cite{ledig2017photo}.
Perceptual loss is previously defined on the activation layers of a pre-trained deep network, where the distance  
between two activated features is minimized. Contrary to the convention, we propose to use features before the 
activation layers, which will overcome two drawbacks of the original design.
First, the activated features are very sparse, especially after a very deep network, as depicted in 
Fig.~\ref{fig:vgg_sparse}.
For example, the average percentage of activated neurons for image `baboon' after VGG19-54\footnote{We use pre-trained 
	19-layer VGG network\cite{simonyan2014very}, where 54 indicates features obtained by the $4^{th}$ convolution 
	before 
	the $5^{th}$ maxpooling layer, representing high-level features and similarly, 22 represents low-level features.} 
	layer 
is merely 11.17\%.
The sparse activation provides weak supervision and thus leads to inferior performance.
Second, using features after activation also causes inconsistent reconstructed brightness compared with the 
ground-truth image, which we will show in Sec.~\ref{subsec:ablation_study}.

Therefore, the total loss for the generator is:
\begin{equation}\label{equ:loss_function}
L_G = L_{\text{percep}} + \lambda{L_G^{Ra}}+ \eta{L_1},
\end{equation}
where $L_1 = \mathbb{E}_{x_i}||G(x_i) - y||_1$ is the content loss that evaluate the 1-norm distance between recovered 
image $G(x_i)$ and the ground-truth $y$, and $\lambda, \eta$ are the coefficients to balance different loss terms.

We also explore a variant of perceptual loss in the PIRM-SR Challenge. 
In contrast to the commonly used perceptual loss that adopts a VGG network trained for image classification, we develop 
a more suitable perceptual loss for SR -- MINC loss.
It is based on a fine-tuned VGG network for material recognition~\cite{bell2015material}, which focuses on textures 
rather than object.
Although the gain of perceptual index brought by MINC loss is marginal, we still believe that exploring perceptual loss 
that focuses on texture is critical for SR.




\subsection{Network Interpolation}
\label{subsec:net_interp}

To remove unpleasant noise in GAN-based methods while maintain a good perceptual quality, we propose a flexible  
and effective strategy -- network interpolation. Specifically, we first train a PSNR-oriented network $G_{\text{PSNR}}$ 
and then obtain a GAN-based network $G_{\text{GAN}}$ by fine-tuning. We interpolate all the corresponding parameters of 
these two networks to derive an interpolated model $G_\text{INTERP}$, whose parameters are:
\begin{equation}
\theta_G^{\text{INTERP}} = (1-\alpha) \ \theta_G^{\text{PSNR}} + \alpha \ \theta_G^{\text{GAN}},
\end{equation}
where $\theta_G^{\text{INTERP}}$, $\theta_G^{\text{PSNR}}$ and $\theta_G^{\text{GAN}}$ are the parameters of  
$G_\text{INTERP}$, $G_\text{PSNR}$ and $G_\text{GAN}$, respectively, and $\alpha \in [0,1]$ is the interpolation 
parameter.

The proposed network interpolation enjoys two merits. First, the interpolated model is able to produce meaningful  
results for any feasible $\alpha$ without introducing artifacts. Second, we can continuously balance perceptual quality 
and fidelity without re-training the model.

We also explore alternative methods to balance the effects of PSNR-oriented and GAN-based methods.
For instance, one can directly interpolate their output images (pixel by pixel) rather than the network parameters.
However, such an approach fails to achieve a good trade-off between noise and blur, i.e., the interpolated image is 
either too blurry or noisy with artifacts (see Sec.~\ref{subsec:exp_net_interp}). 
Another method is to tune the weights of content loss and adversarial loss, i.e., the parameter $\lambda$ and $\eta$ in 
Eq.~(\ref{equ:loss_function}).
But this approach requires tuning loss weights and fine-tuning the network, and thus it is too costly to achieve 
continuous control of the image style.

\section{Experiments} \label{sec:exp}
\subsection{Training Details}

Following SRGAN~\cite{ledig2017photo}, all experiments are performed with a scaling factor of $\times 4$ between LR and 
HR images.
We obtain LR images by down-sampling HR images using the MATLAB bicubic kernel function. The mini-batch size is set to 
16. The spatial size of cropped HR patch is $128 \times 128$.
We observe that training a deeper network benefits from a larger patch size, since an enlarged receptive field helps to 
capture more semantic information. However, it costs more training time and consumes more computing resources. 
This phenomenon is also observed in PSNR-oriented methods (see \textit{supplementary material}).

The training process is divided into two stages. 
First, we train a PSNR-oriented model with the L1 loss. 
The learning rate is initialized as $2\times 10^{-4}$ and decayed by a factor of 2 every $2 \times 10^5$ of mini-batch 
updates. 
We then employ the trained PSNR-oriented model as an initialization for the generator.
The generator is trained using the loss function in Eq.~(\ref{equ:loss_function}) with $\lambda=5 \times 10^{-3}$ 
and $\eta=1 \times 10^{-2}$. 
The learning rate is set to $1\times 10^{-4}$ and halved at [$50k, 100k, 200k, 300k$] iterations.
Pre-training with pixel-wise loss helps GAN-based methods to obtain more visually pleasing results.
The reasons are that 1) it can avoid undesired local optima for the generator; 
2) after pre-training, the discriminator receives relatively good super-resolved images instead of extreme fake ones 
(black or noisy images) at the very beginning, which helps it to focus more on texture discrimination.

For optimization, we use Adam~\cite{kingma2014adam} with $\beta_1=0.9, \beta_2=0.999$. 
We alternately update the generator and discriminator network until the model converges.
We use two settings for our generator -- one of them contains 16 residual blocks, with a capacity similar to that of 
SRGAN and the
other is a deeper model with 23 RRDB blocks.
We implement our models with the PyTorch framework and train them using NVIDIA Titan Xp GPUs.

\subsection{Data}
For training, we mainly use the DIV2K dataset~\cite{agustsson2017ntire}, which is a high-quality (2K resolution) 
dataset 
for image restoration tasks. 
Beyond the training set of DIV2K that contains 800 images, we also seek for other datasets with rich and diverse 
textures for our training.
To this end, we further use the Flickr2K dataset~\cite{timofte2017ntire} consisting of 
2650 2K high-resolution images collected on the Flickr website, and the OutdoorSceneTraining 
(OST)~\cite{wang2018sftgan} 
dataset to enrich our training set. 
We empirically find that using this large dataset with richer textures helps the generator to produce more natural 
results, as shown in Fig.~\ref{fig:ablation}.

We train our models in RGB channels and augment the training dataset with random horizontal flips and 90 degree 
rotations.
We evaluate our models on widely used benchmark datasets -- Set5~\cite{bevilacqua2012low}, 
Set14~\cite{zeyde2010single}, BSD100~\cite{martin2001database}, Urban100~\cite{huang2015single}, and the
PIRM self-validation dataset that is provided in the PIRM-SR Challenge.

\subsection{Qualitative Results}

\begin{figure}[htbp]
	\begin{center}
		\includegraphics[width=0.95\linewidth]{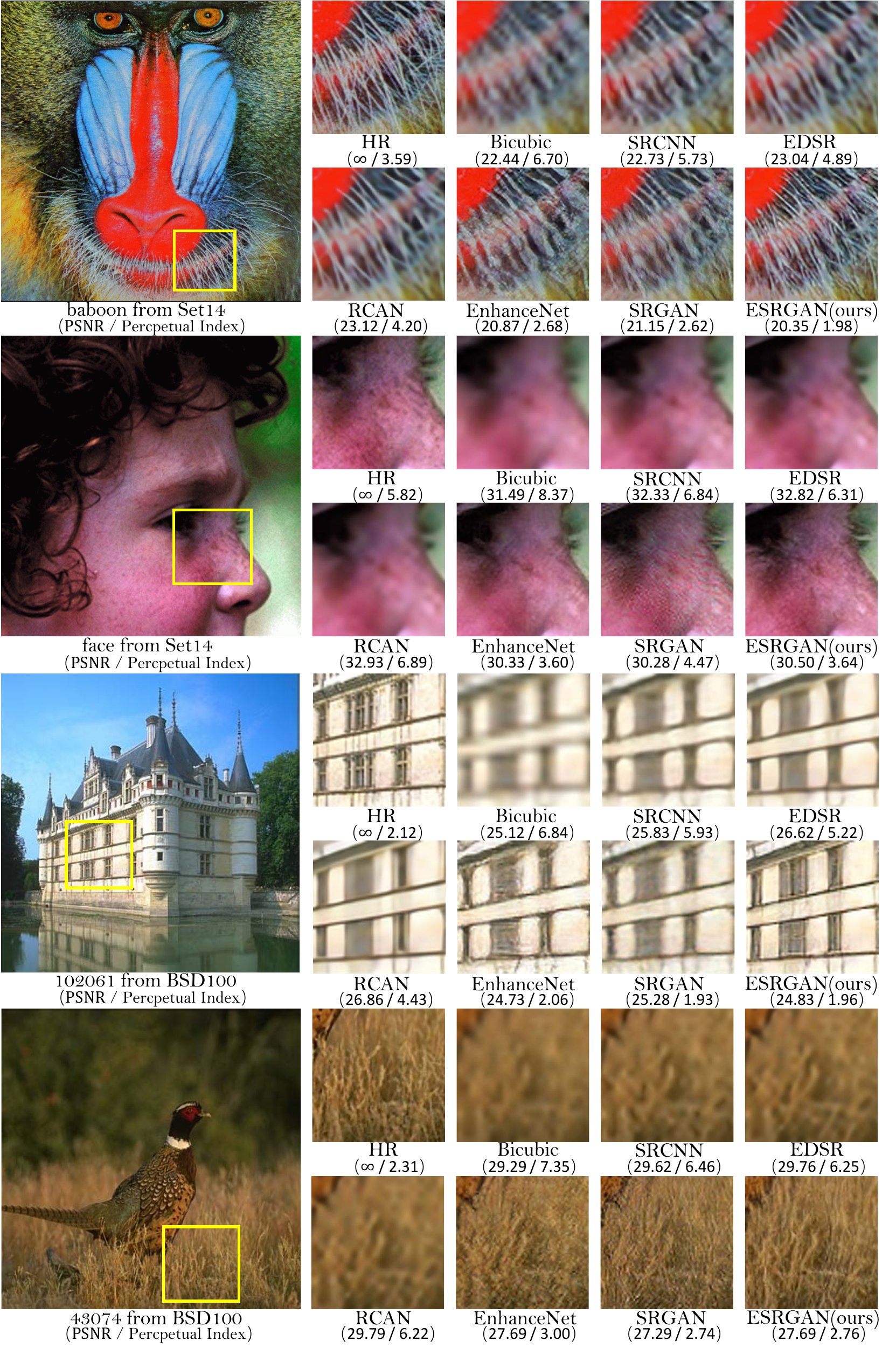}
	\end{center}
	\vspace{-0.5cm}
	\caption{Qualitative results of ESRGAN. ESRGAN produces more natural textures, e.g., animal fur, building structure 
		and grass texture, and also less unpleasant artifacts, e.g., artifacts in the face by SRGAN.}
	\label{fig:qualitative_results}
	\vspace{-0.3cm}
\end{figure}

We compare our final models on several public benchmark datasets with state-of-the-art PSNR-oriented methods including 
SRCNN~\cite{dong2014learning}, EDSR~\cite{lim2017enhanced} and RCAN~\cite{zhang2018image}, and also with 
perceptual-driven approaches including SRGAN~\cite{ledig2017photo} and EnhanceNet \cite{sajjadi2017enhancenet}. 
Since there is no effective and standard metric for perceptual quality, we present some representative qualitative  
results in Fig.~\ref{fig:qualitative_results}. 
PSNR (evaluated on the luminance channel in YCbCr color space) and the perceptual index used in the PIRM-SR Challenge 
are 
also provided for reference.

It can be observed from Fig.~\ref{fig:qualitative_results} that our proposed ESRGAN outperforms previous approaches in 
both sharpness and details.
For instance, ESRGAN can produce sharper and more natural baboon's whiskers and grass textures (see image 43074) than 
PSNR-oriented 
methods, which tend to generate blurry results, and than previous GAN-based methods, whose textures are unnatural and 
contain unpleasing noise. 
ESRGAN is capable of generating more detailed structures in building (see image 102061) while other methods either fail 
to produce enough details (SRGAN) or add undesired textures (EnhanceNet).
Moreover, previous GAN-based methods sometimes introduce unpleasant artifacts, e.g., SRGAN adds wrinkles to the face. 
Our 
ESRGAN gets rid of these artifacts and produces natural results.

\subsection{Ablation Study} \label{subsec:ablation_study}

\begin{figure}[htbp]
	\begin{center}
		\includegraphics[width=\linewidth]{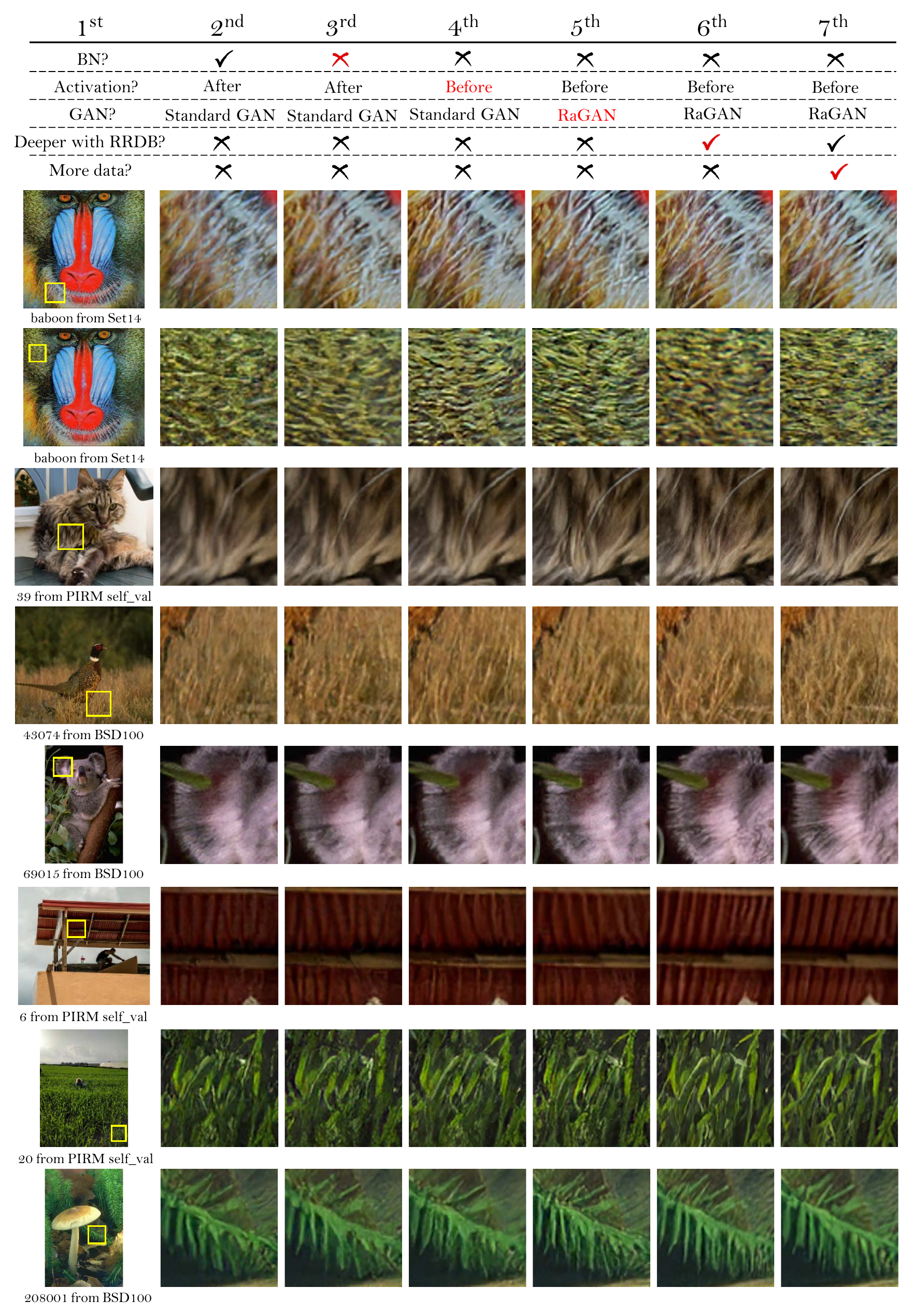}
	\end{center}
	\vspace{-0.3cm}
	\caption{Overall visual comparisons for showing the effects of each component in ESRGAN. Each column represents a 
		model with its configurations in the top. The red sign indicates the main improvement compared with the 
		previous 
		model.}
	\label{fig:ablation}
	\vspace{-0.3cm}
\end{figure}

In order to study the effects of each component in the proposed ESRGAN, we gradually modify the baseline SRGAN model 
and compare their differences. 
The overall visual comparison is illustrated in Fig.~\ref{fig:ablation}. Each column represents a model with its 
configurations shown in the top. The red sign indicates the main improvement compared with the previous model.
A detailed discussion is provided as follows.

\noindent \textbf{BN removal}.
We first remove all BN layers for stable and consistent performance without artifacts.
It does not decrease the performance but saves the computational resources and memory usage.
For some cases, a slight improvement can be observed from the $2^{nd}$ and $3^{rd}$ columns in Fig.~\ref{fig:ablation} 
(e.g., image 39).
Furthermore, we observe that when a network is deeper and more complicated, the model with BN layers is more likely  
to introduce unpleasant artifacts. The examples can be found in the \textit{supplementary material}.

\begin{figure}[htbp]%
	\centering
	\subfloat[brightness influence
	]{{\label{fig:before_activation_a}\includegraphics[width=.5\linewidth]{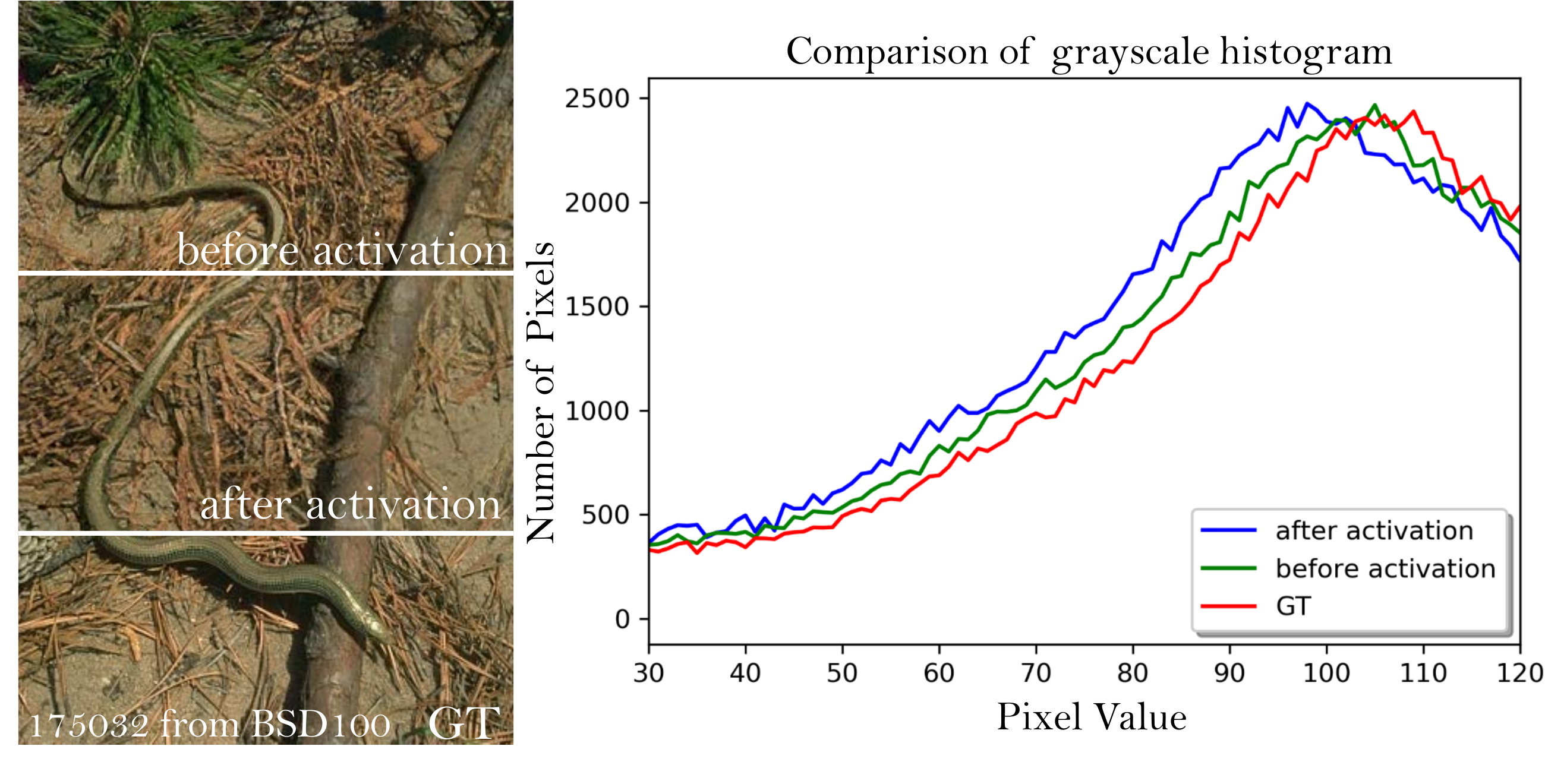}}}%
	\qquad
	\subfloat[detail influence 
	]{{\label{fig:before_activation_b}\includegraphics[width=.42\linewidth]{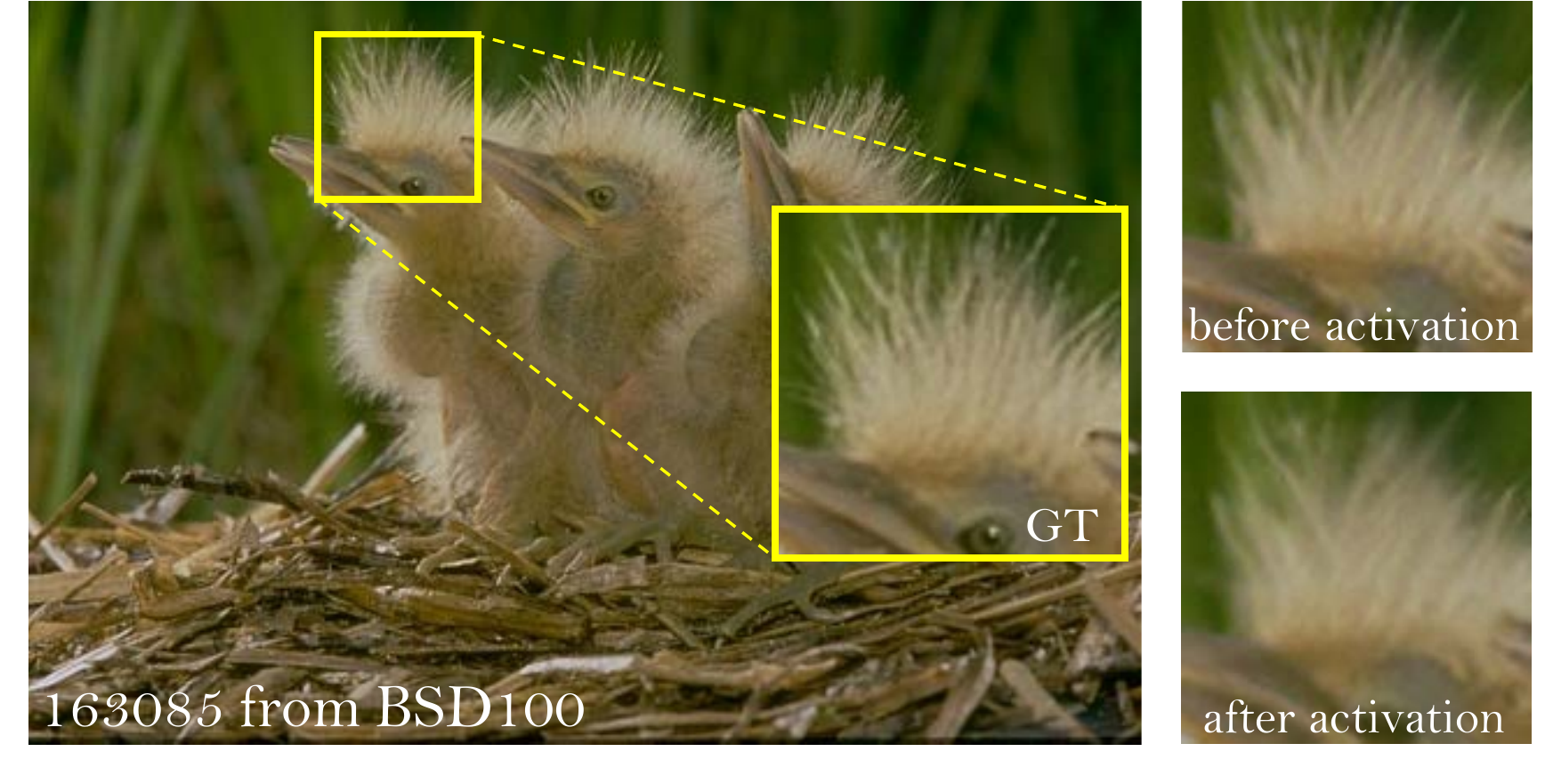}}}%
	\caption{Comparison between before activation and after activation.}%
	\label{fig:before_activation}%
	\vspace{-0.4cm}
\end{figure}

\noindent \textbf{Before activation in perceptual loss}.
We first demonstrate that using features before activation can result in more accurate brightness of reconstructed 
images.
To eliminate the influences of textures and color, we filter the image with a Gaussian kernel and plot the 
histogram of its gray-scale counterpart. 
Fig.~\ref{fig:before_activation_a} shows the distribution of each brightness value.
Using activated features skews the distribution to the left, resulting in a dimmer output while using 
features before activation leads to a more accurate brightness distribution closer to that of the ground-truth.


We can further observe that using features before activation helps to produce sharper edges and richer textures as 
shown in Fig.~\ref{fig:before_activation_b} (see bird feather) and Fig.~\ref{fig:ablation} (see the $3^{rd}$ and 
$4^{th}$ columns), since the dense features before activation offer a stronger supervision than that a sparse 
activation could provide.

\noindent \textbf{RaGAN}. RaGAN uses an improved relativistic discriminator, which is shown to benefit learning 
sharper edges and more detailed textures. 
For example, in the $5^{th}$ column of Fig.~\ref{fig:ablation}, the generated images are sharper with richer textures 
than those on their left (see the baboon, image 39 and image 43074).

\noindent \textbf{Deeper network with RRDB}.
Deeper model with the proposed RRDB can further improve the recovered textures, especially for the regular structures  
like the roof of image 6 in Fig.~\ref{fig:ablation}, since the deep model has a strong representation capacity to 
capture semantic information. 
Also, we find that a deeper model can reduce unpleasing noises like image 20 in Fig.~\ref{fig:ablation}.

In contrast to SRGAN, which claimed that deeper models are increasingly difficult to train, our deeper model  
shows its superior performance with easy training, thanks to the improvements mentioned above especially the proposed 
RRDB without BN layers.

\subsection{Network Interpolation} \label{subsec:exp_net_interp}

We compare the effects of network interpolation and image interpolation strategies in balancing the results of a 
PSNR-oriented model and GAN-based method.
\begin{figure}[htb]
	\begin{center}
		\includegraphics[width=\linewidth]{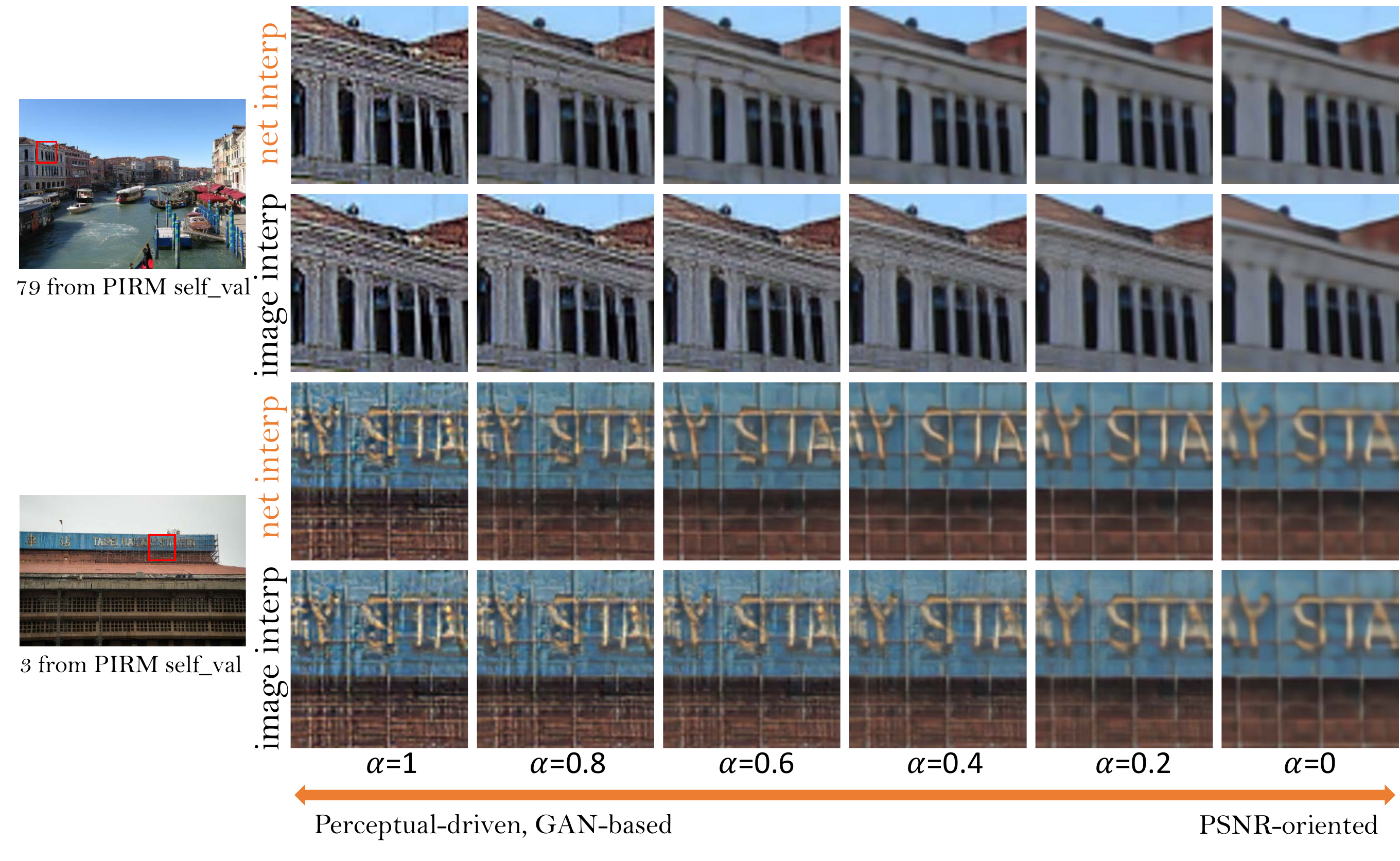}
	\end{center}
	\vspace{-0.4cm}
	\caption{The comparison between network interpolation and image interpolation.}
	\label{fig:net_interp}
	\vspace{-0.4cm}
\end{figure}
We apply simple linear interpolation on both the schemes.
The interpolation parameter $\alpha$ is chosen from 0 to 1 with an interval of 0.2.

As depicted in Fig.~\ref{fig:net_interp}, the pure GAN-based method produces sharp edges and richer textures but with 
some unpleasant artifacts, while the pure PSNR-oriented method outputs cartoon-style blurry images.
By employing network interpolation, unpleasing artifacts are reduced while the textures are maintained. 
By contrast, image interpolation fails to remove these artifacts effectively.

Interestingly, it is observed that the network interpolation strategy provides a smooth control of balancing 
perceptual quality and fidelity in Fig.~\ref{fig:net_interp}.

\subsection{The PIRM-SR Challenge} \label{sec:prim18}
We take a variant of ESRGAN to participate in the PIRM-SR Challenge~\cite{pirm18url}.
Specifically, we use the proposed ESRGAN with 16 residual blocks and also empirically make some modifications to cater 
to the perceptual index.
1) The MINC loss is used as a variant of perceptual loss, as discussed in Sec.~\ref{subsec:perceptual_loss}. 
Despite the marginal gain on the perceptual index, we still believe that exploring perceptual loss that focuses on 
texture is 
crucial for SR.
2) Pristine dataset~\cite{mittal2013making}, which is used for learning the perceptual index, is also employed in our 
training;
3) a high weight of loss $L_1$ up to $\eta=10$ is used due to the PSNR constraints;
4) we also use back projection~\cite{timofte2016seven} as post-processing, which can improve PSNR and sometimes 
lower the perceptual index.

For other regions 1 and 2 that require a higher PSNR, we use image interpolation between the results of our ESRGAN and  
those of a PSNR-oriented method RCAN~\cite{zhang2018image}. 
The image interpolation scheme achieves a lower perceptual index (lower is better) although we observed more visually 
pleasing results by using the network interpolation scheme.
%
%
Our proposed ESRGAN model won the first place in the PIRM-SR Challenge (region 3) with the best perceptual index.
%
%

\section{Conclusion}

We have presented an ESRGAN model that achieves consistently better perceptual quality than previous SR 
methods. The method won the first place in the PIRM-SR Challenge in terms of the perceptual index. 
We have formulated a novel architecture containing several RDDB blocks without BN layers. In addition, useful 
techniques including residual scaling and smaller initialization are employed to facilitate the training of the 
proposed deep model.
We have also introduced the use of relativistic GAN as the discriminator, which learns to judge whether one image is 
more realistic  
than another, guiding the generator to recover more detailed textures.
Moreover, we have enhanced the perceptual loss by using the features before activation, which offer stronger 
supervision 
and thus restore more accurate brightness and realistic textures.


\vspace{0.5cm}
\noindent\textbf{Acknowledgement}.
This work is supported by SenseTime Group Limited, the General Research Fund sponsored by the Research Grants 
Council of the Hong Kong SAR (CUHK 14241716, 14224316. 14209217), National Natural Science Foundation of China 
(U1613211) and Shenzhen Research Program \\ (JCYJ20170818164704758, JCYJ20150925163005055).


{\scriptsize 
\bibliographystyle{splncs}
\bibliography{short,bib}

\begin{thebibliography}{10}

\bibitem{ledig2017photo}
Ledig, C., Theis, L., Husz{\'a}r, F., Caballero, J., Cunningham, A., Acosta,
  A., Aitken, A., Tejani, A., Totz, J., Wang, Z.,  et~al.:
\newblock Photo-realistic single image super-resolution using a generative
  adversarial network.
\newblock In: CVPR. (2017)

\bibitem{jolicoeur2018relativistic}
Jolicoeur-Martineau, A.:
\newblock The relativistic discriminator: a key element missing from standard
  gan.
\newblock arXiv preprint arXiv:1807.00734 (2018)

\bibitem{pirm18url}
Blau, Y., Mechrez, R., Timofte, R., Michaeli, T., Zelnik-Manor, L.:
\newblock The pirm challenge on perceptual super resolution.
\newblock \url{https://www.pirm2018.org/PIRM-SR.html} (2018)

\bibitem{dong2014learning}
Dong, C., Loy, C.C., He, K., Tang, X.:
\newblock Learning a deep convolutional network for image super-resolution.
\newblock In: ECCV. (2014)

\bibitem{kim2016accurate}
Kim, J., Kwon~Lee, J., Mu~Lee, K.:
\newblock Accurate image super-resolution using very deep convolutional
  networks.
\newblock In: CVPR. (2016)

\bibitem{lai2017deep}
Lai, W.S., Huang, J.B., Ahuja, N., Yang, M.H.:
\newblock Deep laplacian pyramid networks for fast and accurate
  super-resolution.
\newblock In: CVPR. (2017)

\bibitem{kim2016deeply}
Kim, J., Kwon~Lee, J., Mu~Lee, K.:
\newblock Deeply-recursive convolutional network for image super-resolution.
\newblock In: CVPR. (2016)

\bibitem{tai2017image}
Tai, Y., Yang, J., Liu, X.:
\newblock Image super-resolution via deep recursive residual network.
\newblock In: CVPR. (2017)

\bibitem{tai2017memnet}
Tai, Y., Yang, J., Liu, X., Xu, C.:
\newblock Memnet: A persistent memory network for image restoration.
\newblock In: ICCV. (2017)

\bibitem{haris2018deep}
Haris, M., Shakhnarovich, G., Ukita, N.:
\newblock Deep backprojection networks for super-resolution.
\newblock In: CVPR. (2018)

\bibitem{zhang2018residual}
Zhang, Y., Tian, Y., Kong, Y., Zhong, B., Fu, Y.:
\newblock Residual dense network for image super-resolution.
\newblock In: CVPR. (2018)

\bibitem{zhang2018image}
Zhang, Y., Li, K., Li, K., Wang, L., Zhong, B., Fu, Y.:
\newblock Image super-resolution using very deep residual channel attention
  networks.
\newblock In: ECCV. (2018)

\bibitem{johnson2016perceptual}
Johnson, J., Alahi, A., Fei-Fei, L.:
\newblock Perceptual losses for real-time style transfer and super-resolution.
\newblock In: ECCV. (2016)

\bibitem{bruna2015super}
Bruna, J., Sprechmann, P., LeCun, Y.:
\newblock Super-resolution with deep convolutional sufficient statistics.
\newblock In: ICLR. (2015)

\bibitem{goodfellow2014generative}
Goodfellow, I., Pouget-Abadie, J., Mirza, M., Xu, B., Warde-Farley, D., Ozair,
  S., Courville, A., Bengio, Y.:
\newblock Generative adversarial nets.
\newblock In: NIPS. (2014)

\bibitem{sajjadi2017enhancenet}
Sajjadi, M.S., Sch{\"o}lkopf, B., Hirsch, M.:
\newblock Enhancenet: Single image super-resolution through automated texture
  synthesis.
\newblock In: ICCV. (2017)

\bibitem{wang2018sftgan}
Wang, X., Yu, K., Dong, C., Loy, C.C.:
\newblock Recovering realistic texture in image super-resolution by deep
  spatial feature transform.
\newblock In: CVPR. (2018)

\bibitem{he2016deep}
He, K., Zhang, X., Ren, S., Sun, J.:
\newblock Deep residual learning for image recognition.
\newblock In: CVPR. (2016)

\bibitem{ioffe2015batch}
Ioffe, S., Szegedy, C.:
\newblock Batch normalization: Accelerating deep network training by reducing
  internal covariate shift.
\newblock In: ICMR. (2015)

\bibitem{lim2017enhanced}
Lim, B., Son, S., Kim, H., Nah, S., Lee, K.M.:
\newblock Enhanced deep residual networks for single image super-resolution.
\newblock In: CVPRW. (2017)

\bibitem{szegedy2016inception}
Szegedy, C., Ioffe, S., Vanhoucke, V.:
\newblock Inception-v4, inception-resnet and the impact of residual connections
  on learning.
\newblock arXiv preprint arXiv:1602.07261 (2016)

\bibitem{blau2017perception}
Blau, Y., Michaeli, T.:
\newblock The perception-distortion tradeoff.
\newblock In: CVPR. (2017)

\bibitem{ma2017learning}
Ma, C., Yang, C.Y., Yang, X., Yang, M.H.:
\newblock Learning a no-reference quality metric for single-image
  super-resolution.
\newblock CVIU \textbf{158} (2017)  1--16

\bibitem{mittal2013making}
Mittal, A., Soundararajan, R., Bovik, A.C.:
\newblock Making a completely blind image quality analyzer.
\newblock IEEE Signal Process. Lett. \textbf{20}(3) (2013)  209--212

\bibitem{dong2016image}
Dong, C., Loy, C.C., He, K., Tang, X.:
\newblock Image super-resolution using deep convolutional networks.
\newblock TPAMI \textbf{38}(2) (2016)  295--307

\bibitem{yu2018crafting}
Yu, K., Dong, C., Lin, L., Loy, C.C.:
\newblock Crafting a toolchain for image restoration by deep reinforcement
  learning.
\newblock In: CVPR. (2018)

\bibitem{yuan2018unsupervised}
Yuan, Y., Liu, S., Zhang, J., Zhang, Y., Dong, C., Lin, L.:
\newblock Unsupervised image super-resolution using cycle-in-cycle generative
  adversarial networks.
\newblock In: CVPRW. (2018)

\bibitem{he2015delving}
He, K., Zhang, X., Ren, S., Sun, J.:
\newblock Delving deep into rectifiers: Surpassing human-level performance on
  imagenet classification.
\newblock In: ICCV. (2015)

\bibitem{gatys2015texture}
Gatys, L., Ecker, A.S., Bethge, M.:
\newblock Texture synthesis using convolutional neural networks.
\newblock In: NIPS. (2015)

\bibitem{roey2018maintaining}
Mechrez, R., Talmi, I., Shama, F., Zelnik-Manor, L.:
\newblock Maintaining natural image statistics with the contextual loss.
\newblock arXiv preprint arXiv:1803.04626 (2018)

\bibitem{arjovsky2017wasserstein}
Arjovsky, M., Chintala, S., Bottou, L.:
\newblock Wasserstein gan.
\newblock arXiv preprint arXiv:1701.07875 (2017)

\bibitem{gulrajani2017improved}
Gulrajani, I., Ahmed, F., Arjovsky, M., Dumoulin, V., Courville, A.C.:
\newblock Improved training of wasserstein gans.
\newblock In: NIPS. (2017)

\bibitem{miyato2018spectral}
Miyato, T., Kataoka, T., Koyama, M., Yoshida, Y.:
\newblock Spectral normalization for generative adversarial networks.
\newblock arXiv preprint arXiv:1802.05957 (2018)

\bibitem{huang2016densely}
Huang, G., Liu, Z., Weinberger, K.Q., van~der Maaten, L.:
\newblock Densely connected convolutional networks.
\newblock In: CVPR. (2017)

\bibitem{nah2017deep}
Nah, S., Kim, T.H., Lee, K.M.:
\newblock Deep multi-scale convolutional neural network for dynamic scene
  deblurring.
\newblock In: CVPR. (2017)

\bibitem{zhang2017residual}
Zhang, K., Sun, M., Han, X., Yuan, X., Guo, L., Liu, T.:
\newblock Residual networks of residual networks: Multilevel residual networks.
\newblock IEEE Transactions on Circuits and Systems for Video Technology (2017)

\bibitem{simonyan2014very}
Simonyan, K., Zisserman, A.:
\newblock Very deep convolutional networks for large-scale image recognition.
\newblock arXiv preprint arXiv:1409.1556 (2014)

\bibitem{bell2015material}
Bell, S., Upchurch, P., Snavely, N., Bala, K.:
\newblock Material recognition in the wild with the materials in context
  database.
\newblock In: CVPR. (2015)

\bibitem{kingma2014adam}
Kingma, D., Ba, J.:
\newblock Adam: A method for stochastic optimization.
\newblock In: ICLR. (2015)

\bibitem{agustsson2017ntire}
Agustsson, E., Timofte, R.:
\newblock Ntire 2017 challenge on single image super-resolution: Dataset and
  study.
\newblock In: CVPRW. (2017)

\bibitem{timofte2017ntire}
Timofte, R., Agustsson, E., Van~Gool, L., Yang, M.H., Zhang, L., Lim, B., Son,
  S., Kim, H., Nah, S., Lee, K.M.,  et~al.:
\newblock Ntire 2017 challenge on single image super-resolution: Methods and
  results.
\newblock In: CVPRW. (2017)

\bibitem{bevilacqua2012low}
Bevilacqua, M., Roumy, A., Guillemot, C., Alberi-Morel, M.L.:
\newblock Low-complexity single-image super-resolution based on nonnegative
  neighbor embedding.
\newblock In: BMVC, BMVA press (2012)

\bibitem{zeyde2010single}
Zeyde, R., Elad, M., Protter, M.:
\newblock On single image scale-up using sparse-representations.
\newblock In: International Conference on Curves and Surfaces, Springer (2010)

\bibitem{martin2001database}
Martin, D., Fowlkes, C., Tal, D., Malik, J.:
\newblock A database of human segmented natural images and its application to
  evaluating segmentation algorithms and measuring ecological statistics.
\newblock In: ICCV. (2001)

\bibitem{huang2015single}
Huang, J.B., Singh, A., Ahuja, N.:
\newblock Single image super-resolution from transformed self-exemplars.
\newblock In: CVPR. (2015)

\bibitem{timofte2016seven}
Timofte, R., Rothe, R., Van~Gool, L.:
\newblock Seven ways to improve example-based single image super resolution.
\newblock In: CVPR. (2016)

\bibitem{glorot2010understanding}
Glorot, X., Bengio, Y.:
\newblock Understanding the difficulty of training deep feedforward neural
  networks.
\newblock In: International Conference on Artificial Intelligence and
  Statistics. (2010)

\end{thebibliography}
}


\pagestyle{headings}

\title{ESRGAN: Enhanced Super-Resolution Generative Adversarial Networks Supplementary File\vspace{-0.3cm}}

\titlerunning{ESRGAN Supplementary File}

\authorrunning{Xintao Wang \textit{et al.}}

\author{
	Xintao Wang$^1$, Ke Yu$^1$, Shixiang Wu$^2$, Jinjin Gu$^3$, Yihao Liu$^4$, \\
	Chao Dong$^2$, Chen Change Loy$^5$, Yu Qiao$^2$, Xiaoou Tang$^1$
}
\institute{
	\scriptsize
	$^1$CUHK-SenseTime Joint Lab, The Chinese University of Hong Kong \\
	$^2$SIAT-SenseTime Joint Lab, Shenzhen Institutes of Advanced Technology,\\Chinese Academy of Sciences
	$^3$The Chinese University of Hong Kong, Shenzhen \\
	$^4$University of Chinese Academy of Sciences
	$^5$Nanyang Technological University, Singapore \\
	\email{ \{wx016,yk017,xtang\}@ie.cuhk.edu.hk, \{sx.wu,chao.dong,yu.qiao\}@siat.ac.cn 
		liuyihao14@mails.ucas.ac.cn, 115010148@link.cuhk.edu.cn, ccloy@ntu.edu.sg }
}

\maketitle
\vspace{-0.6cm}
\begin{abstract}
	In this supplementary file, we first show more examples of Batch-Normalization (BN) related artifacts in 
	Section~\ref{sec:BN_artifacts}. 
	Then we introduce several useful techniques that facilitate training very deep models in 
	Section~\ref{sec:useful_techniques}.
	The analysis of the influence of different datasets and training patch size is depicted in 
	Section~\ref{sec:influence_datasets} and Section~\ref{sec:influence_patch_size}, respectively. 
	Finally, in Section~\ref{sec:qualitative_cmp}, we provide more qualitative results for visual comparison.
\end{abstract}

\section{BN artifacts} \label{sec:BN_artifacts}

\begin{figure}[htbp]
	\vspace{-0.2cm}
	\begin{center}
		\includegraphics[width=\linewidth]{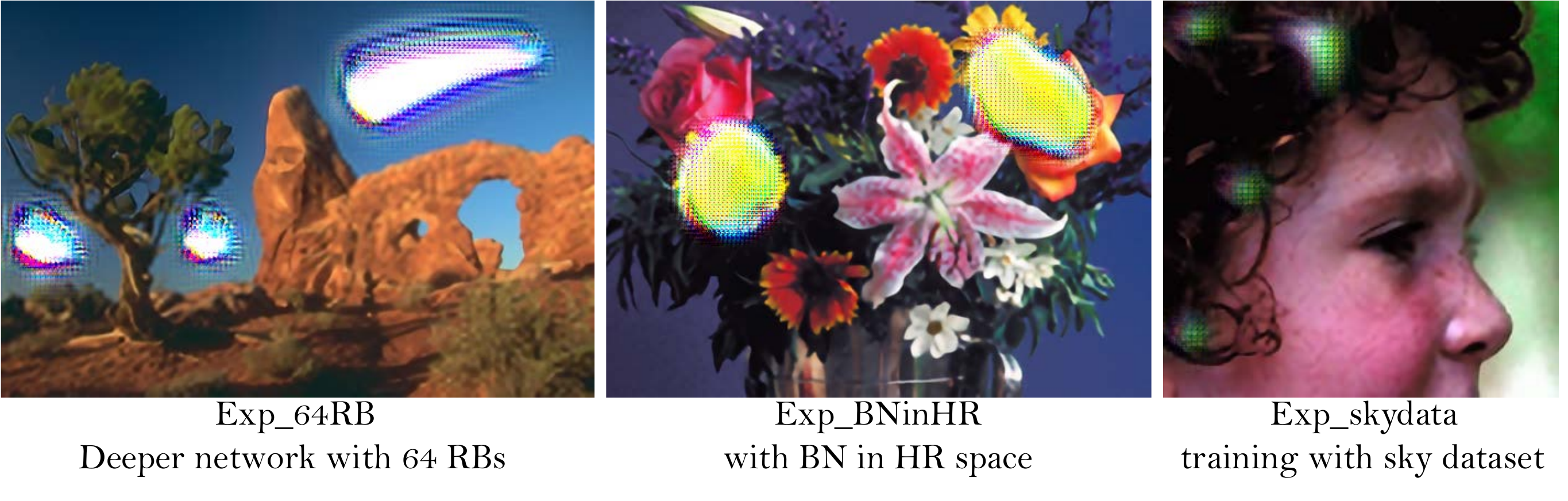}
	\end{center}
	\vspace{-0.4cm}
	\caption{Examples of BN artifacts in PSNR-oriented methods. The BN artifacts are more likely to appear in deeper 
		networks, with BN in HR space and using mismatched dataset whose statistics are different from those of 
		testing dataset.}
	\vspace{-0.4cm}
	\label{fig:BN_artifacts_PSNR}
\end{figure}

\begin{figure}[htbp]
	\vspace{-0.4cm}
	\begin{center}
		\includegraphics[width=\linewidth]{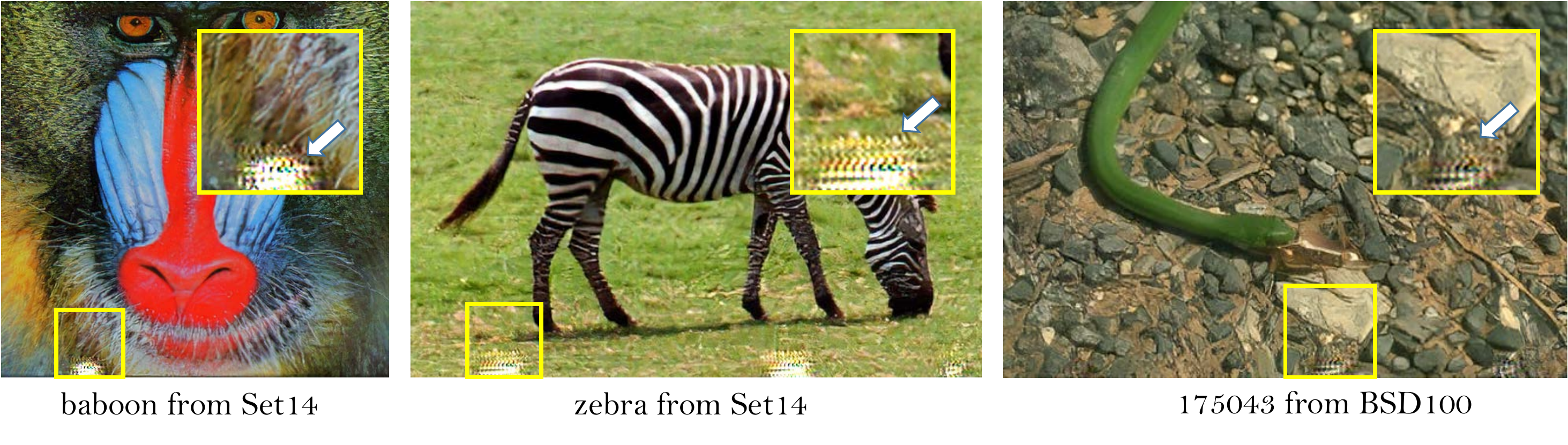}
	\end{center}
	\vspace{-0.4cm}
	\caption{Examples of BN artifacts in models under the GAN framework.}
	\label{fig:BN_artifacts_SRGAN}
	\vspace{-0.4cm}
\end{figure}

\begin{figure}[htbp]
	\vspace{-0.4cm}
	\begin{center}
		\includegraphics[width=\linewidth]{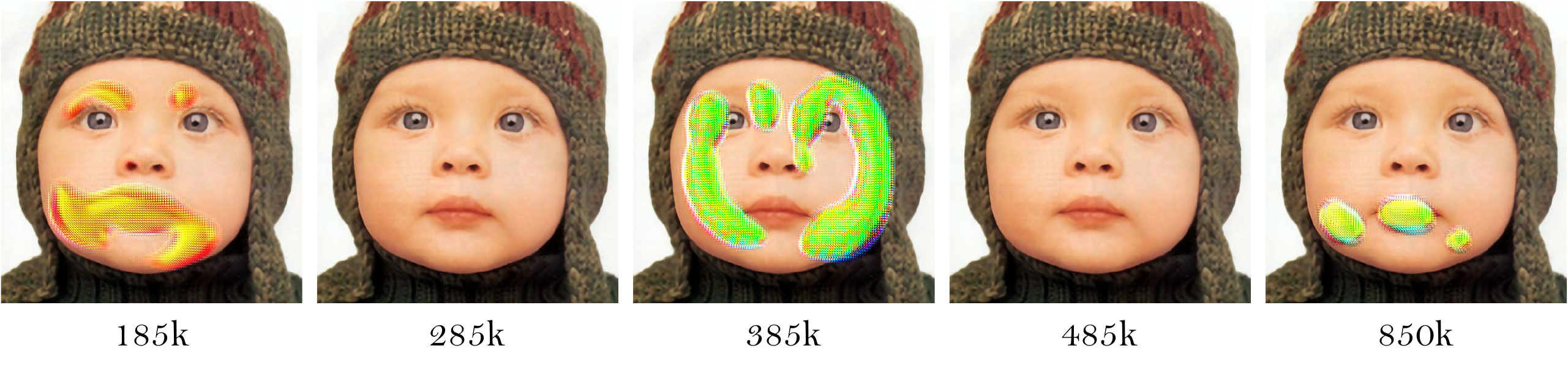}
	\end{center}
	\vspace{-0.4cm}
	\caption{Evolution of the model Exp\_BNinHR (with BN in HR space) during training progress.The BN artifacts 
		occasionally appear over training, resulting in unstable  performance.}
	\vspace{-0.4cm}
	\label{fig:BN_artifacts_training}
\end{figure}

We empirically observe that BN layers tend to bring artifacts.
These artifacts, namely BN artifacts, occasionally appear among iterations and different settings, violating the needs 
for a stable performance over training.
In this section, we present that the network depth, BN position, training dataset and training loss have impact on 
the occurrence of BN artifacts and show corresponding visual examples in 
Fig.~\ref{fig:BN_artifacts_PSNR},~\ref{fig:BN_artifacts_SRGAN} and~\ref{fig:BN_artifacts_training}.

\begin{table}[htbp]
	\vspace{-0.4cm}
	\centering
	\caption{Experimental variants for exploring BN artifacts.}
	\label{tb:BN}
	\begin{tabular}{|c|c|c|c|c|}
		\hline
		Name & Number of RB & BN position & training dataset & training loss \\ \hline
		Exp\_base & 16 & LR space & DIV2K & $L1$ \\ \hline
		Exp\_BNinHR & 16 & LR and HR space & DIV2K & $L1$ \\ \hline
		Exp\_64RB & 64 & LR space & DIV2K & $L1$ \\ \hline
		Exp\_skydata & 16 & LR space & sky data & $L1$ \\ \hline
		Exp\_SRGAN & 16 & LR space & DIV2K & $VGG+GAN+L1$ \\ \hline
	\end{tabular}
	\vspace{-0.4cm}
\end{table}

To explore BN artifacts, we conduct several experiments as shown in Tab.~\ref{tb:BN}.
The baseline is similar to SRResNet~\cite{ledig2017photo} with 16 Residual Blocks (RB) and all the BN layers are in 
the LR space, i.e., before up-sampling layers. 
The baseline setting is unlikely to introduce BN artifacts in our experiments.
However, if the network goes deeper or there is an extra BN layer in HR space (i.e., after up-sampling layers), BN 
artifacts are more likely to appear (see examples in Fig.~\ref{fig:BN_artifacts_PSNR}). 

When we replace the training dataset of the baseline with the sky dataset~\cite{wang2018sftgan}, the BN artifacts 
appear (see examples in Fig.~\ref{fig:BN_artifacts_PSNR}).
BN layers normalize the features using mean and variance in a batch during training while using estimated 
mean and variance of the whole training dataset during testing.
Therefore, when the statistics of training (e.g., sky dataset) and testing datasets differ a lot, BN layers tend to 
introduce unpleasant artifacts and limit the generalization ability.

Training in a GAN framework increases the occurrence probability of BN artifacts in our experiments. 
We employ the same network structure as baseline and replace the $L1$ loss with $VGG+GAN+L1$ loss.
The BN artifacts become more likely to appear and the visual examples are shown in Fig.~\ref{fig:BN_artifacts_SRGAN}.

The BN artifacts occasionally appear over training, i.e, the BN artifacts appear, disappear and change on different 
training iterations, as shown in Fig~\ref{fig:BN_artifacts_training}.
We therefore remove BN layers for stable training and consistent performance.
The reasons behind and potential solutions remain to be further studied.

\section{Useful techniques to train a very deep network} \label{sec:useful_techniques}

Since we remove BN layers for stable training and consistent performance, training a very deep network becomes a 
problem. 
Despite the proposed Residual-in-Residual Dense Block (RRDB), which takes advantages of residual learning and more 
connections, we also find two useful techniques to ease the training of a very deep networks -- smaller initialization 
and 
residual scaling. 

Initialization is important for a very deep network especially without BN 
layers~\cite{glorot2010understanding,he2015delving}.
He et al.~\cite{he2015delving} propose a robust initialization method, namely MSRA initialization, 
that is suitable for VGG-style network (plain network \mbox{without} residual connections). 
The assumption is that a proper initialization method should avoid reducing or magnifying the magnitudes of input 
signals exponentially.
It is worth noting that this assumption no longer holds due to the residual path in ResNet~\cite{he2016deep}, leading 
to a magnified magnitudes of input signals.
This problem is alleviated by normalizing the features with BN layers~\cite{ioffe2015batch}.
For a very deep network containing residual blocks without BN layers, a new initialization method should be applied. 
We find a smaller initialization than MSRA initialization (multiplying 0.1 for all initialization parameters 
that calculated by MSRA initialization) works well in our experiments.

Another method for training deeper networks is residual learning, proposed by Szegedy et 
al.~\cite{szegedy2016inception} and also used in used in EDSR~\cite{lim2017enhanced}.
It scales down the residuals by multiplying a constant between 0 and 1 before adding them to the main path to prevent 
instability. In our settings, for each residual block, the residual features after the last convolution layer are 
multiplied by 0.2.
Intuitively, the residual scaling can be interpreted to correct the improper initialization, thus avoiding magnifying 
the magnitudes of input signals in residual networks.

\begin{figure}[htbp]%
	\vspace{-0.4cm}
	\centering
	\subfloat[
	]{{\label{fig:deeper_net_a}\includegraphics[width=.47\linewidth]{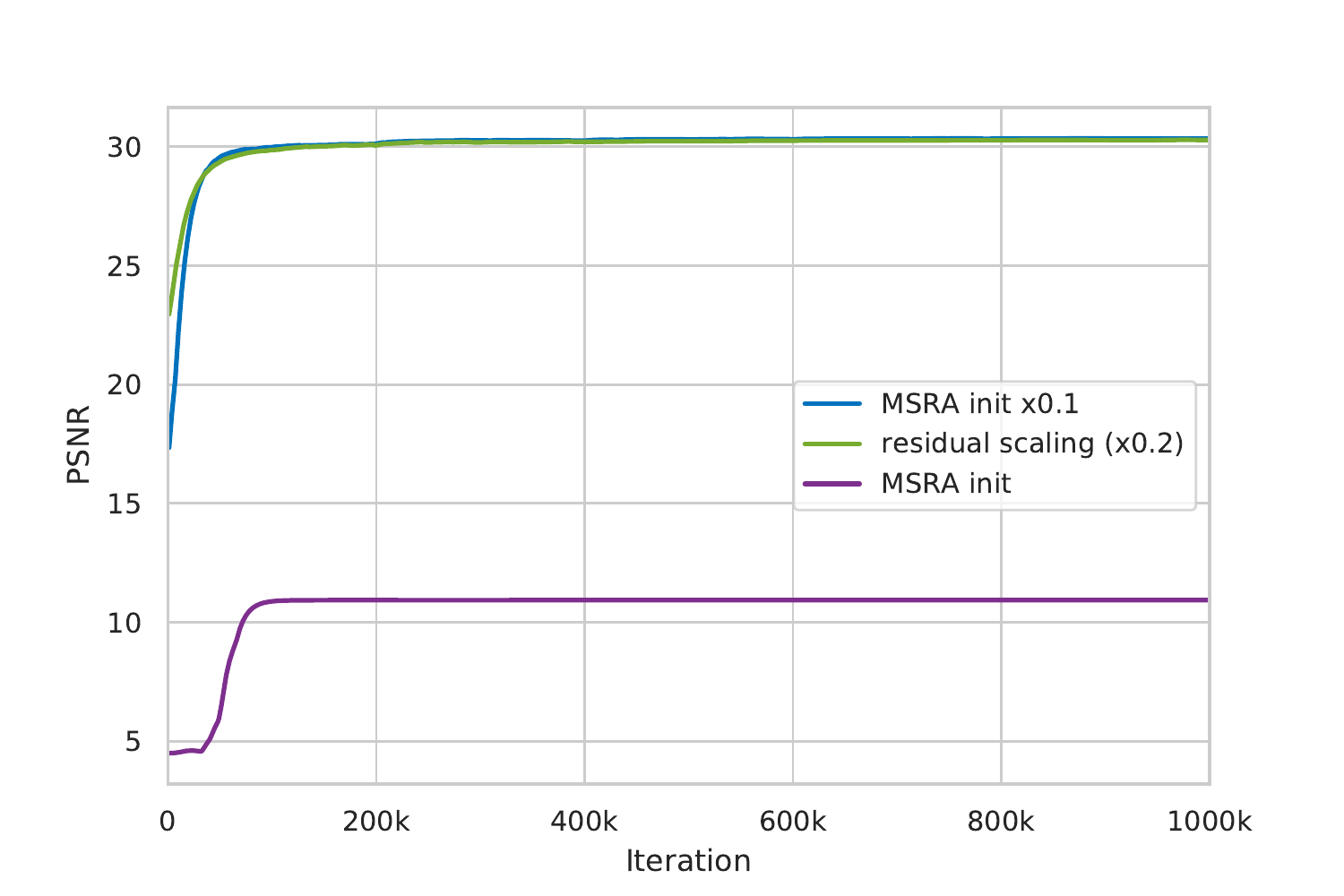}}}%
	\qquad
	\subfloat[ 
	]{{\label{fig:deeper_net_b}\includegraphics[width=.47\linewidth]{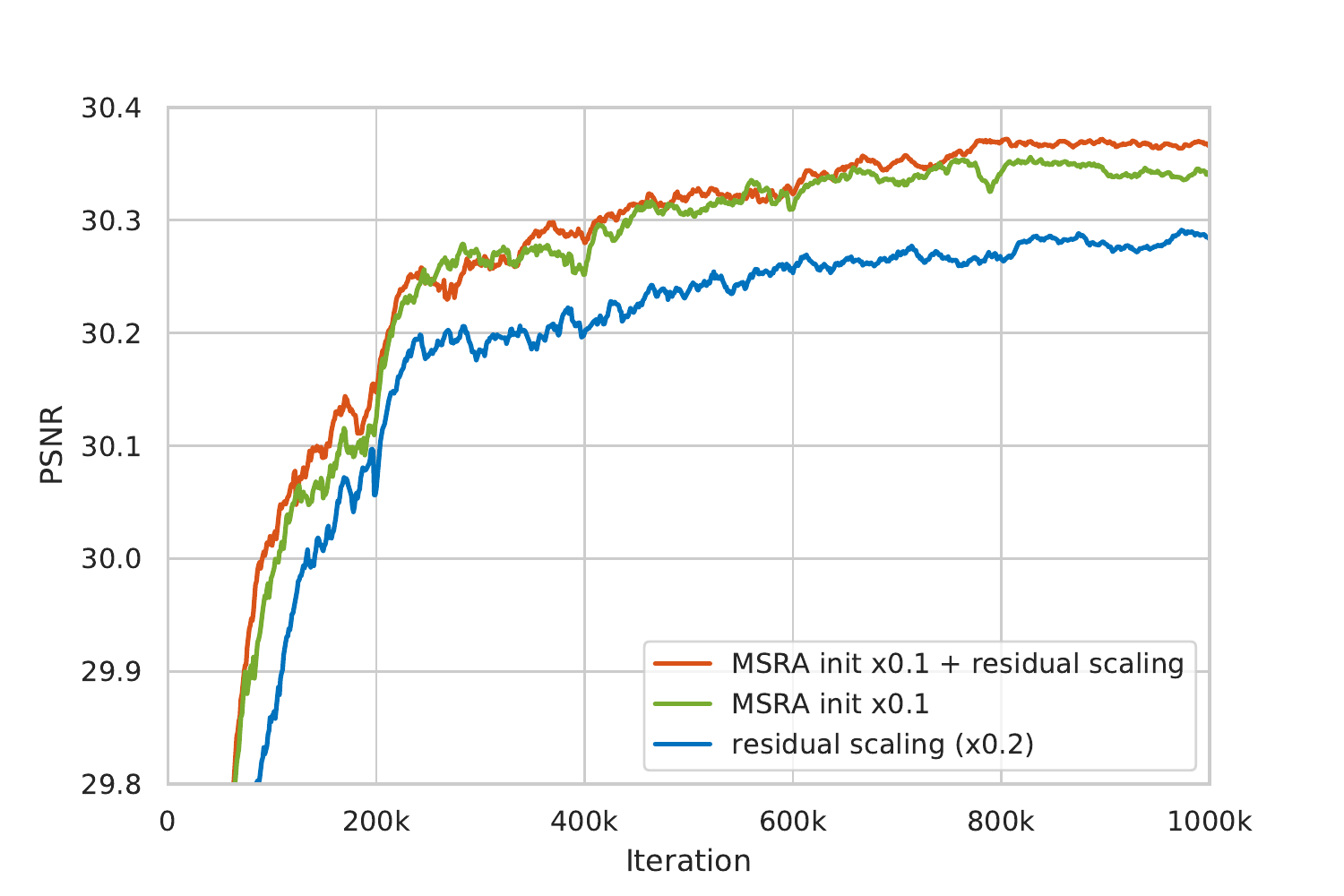}}}%
	\caption{Smaller initialization and residual scaling benefit the convergence and the performance of very deep 
		networks (PSNR is evaluated on Set5 with RGB channels).}%
	\label{fig:deeper_net}%
	\vspace{-0.4cm}
\end{figure}

We use a very deep network containing 64 RBs for experiments. As shown in Fig.~\ref{fig:deeper_net_a}, if we simply 
use MSRA initialization, the network falls into an extremely bad local minimum with poor performance. 
However, smaller initialization ($\times 0.1$) helps the network to jump out the bad local minimum and achieve good 
performance.
The zoomed curves are shown in Fig.~\ref{fig:deeper_net_b}. Smaller initialization achieves a higher PSNR than residual 
scaling. In addition, we can use both techniques to further obtain a slight improvement. 

\section{The influence of different datasets} \label{sec:influence_datasets}

First we show that larger datasets lead to better performance for PSNR-oriented methods.
We use a large model, where 23 Residual-in-Residual Blocks (RRDB) are placed before 
the upsampling layer followed by two convolution layers for reconstruction.
The overall comparison of quantitative evaluation can be found in Tab.~\ref{tb:psnrssim_y}.

A widely used training dataset is DIV2K~\cite{agustsson2017ntire} that contains 800 images.
We also explore other datasets with more diverse scenes -- Flickr2K dataset~\cite{timofte2017ntire} consisting of 
2650 2K high-resolution images collected on the Flickr website.
It is observed that the merged dataset with DIV2K and Flickr2K, namely DF2K dataset, increases the PSNR performance  
 (see Tab.~\ref{tb:psnrssim_y}).

\begin{table}[hbtp]
	\vspace{-0.5cm}
	\scriptsize
	\centering
	\caption{Quantitative evaluation of state-of-the-art PSNR-oriented SR algorithms: average PSNR/SSIM on Y 
		channel. The best and second best results are \textbf{highlighted} and \underline{underlined}, respectively.}
	\label{tb:psnrssim_y}
	\vspace{0.2cm}
	\begin{tabular}{|c|c|c|c|c|c|c|}
		\hline
		\multicolumn{2}{|c|}{\multirow{2}*{\tabincell{c}{Method\\ with training data}}} & Set5 & Set14 & BSD100 & 
		Urban100 & Manga109 \\
		\cline{3-7}
		\multicolumn{2}{|c|}{~} & PSNR/SSIM & PSNR/SSIM & PSNR/SSIM & PSNR/SSIM & PSNR/SSIM \\ 
		\hline
		Bicubic & - & 28.42/0.8104 & 26.00/0.7027 & 25.96/0.6675 & 23.14/0.6577 & 24.89/0.7866 \\ 
		SRCNN~\cite{dong2014learning}& 291 & 30.48/0.8628 & 27.50/0.7513 & 26.90/0.7101 & 24.52/0.7221 & 27.58/0.8555 \\
		MemNet~\cite{tai2017memnet}& 291 & 31.74/0.8893 & 28.26/0.7723 & 27.40/0.7281 & 25.50/0.7630 & 29.42/0.8942 \\
		EDSR~\cite{lim2017enhanced}& DIV2K & 32.46/0.8968 & 28.80/0.7876 & 27.71/0.7420 & 26.64/0.8033 & 31.02/0.9148 \\
		RDN~\cite{zhang2018residual}& DIV2K & 32.47/0.8990 & 28.81/0.7871 & 27.72/0.7419 & 26.61/0.8028 & 31.00/0.9151 
		\\
		RCAN~\cite{zhang2018image}& DIV2K & \underline{32.63}/\underline{0.9002} & 28.87/0.7889 &  
		\underline{27.77}/\underline{0.7436} & \underline{26.82}/ \underline{0.8087} & 
		\underline{31.22}/ \underline{0.9173} \\
		RRDB(ours)& DIV2K  & 32.60/\underline{0.9002} & \underline{28.88}/\underline{0.7896} &  27.76/ 
		0.7432 & 26.73/0.8072 & 31.16/0.9164 \\
		RRDB(ours)& DF2K& \textbf{32.73}/\textbf{0.9011} & \textbf{28.99}/\textbf{0.7917} & 
		\textbf{27.85}/\textbf{0.7455} & \textbf{27.03}/\textbf{0.8153} & \textbf{31.66}/\textbf{0.9196}\\
		\hline 
	\end{tabular}
\end{table}

\begin{figure}[htbp]
	\vspace{-0.4cm}
	\begin{center}
		\includegraphics[width=0.95\linewidth]{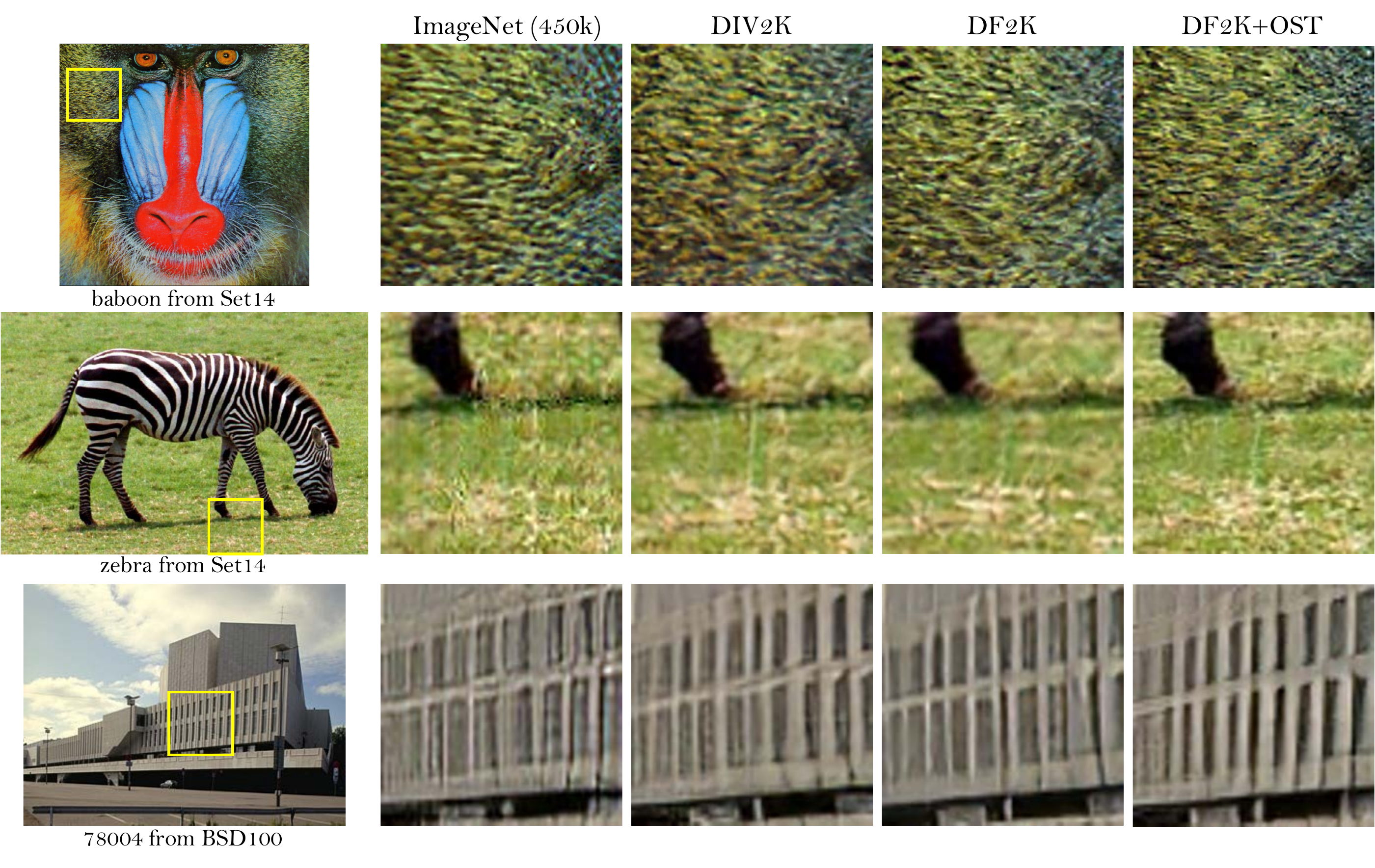}
	\end{center}
	\vspace{-0.4cm}
	\caption{The influence of different datasets.}
	\vspace{-0.5cm}
	\label{fig:data_cmp}
\end{figure}

For perceptual-driven methods that focus on texture restoration, we further enrich the training set with
OutdoorSceneTraining (OST)~\cite{wang2018sftgan} dataset with diverse natural textures.
We employ the large model with 23 RRDB blocks. A subset of ImageNet containing about 
$450k$ images is also used for comparison. The qualitative results are shown in Fig.~\ref{fig:data_cmp}. Training with 
ImageNet introduces new types of artifacts as in image zebra of Fig.~\ref{fig:data_cmp} while OST dataset
benefits the grass restoration. 

\section{The influence of training patch size} \label{sec:influence_patch_size}

\begin{figure}[htbp]%
	\vspace{-0.7cm}
	\centering
	\subfloat[16 Residual Blocks
	]{{\label{fig:patch_size_ResNet}\includegraphics[width=.45\linewidth]{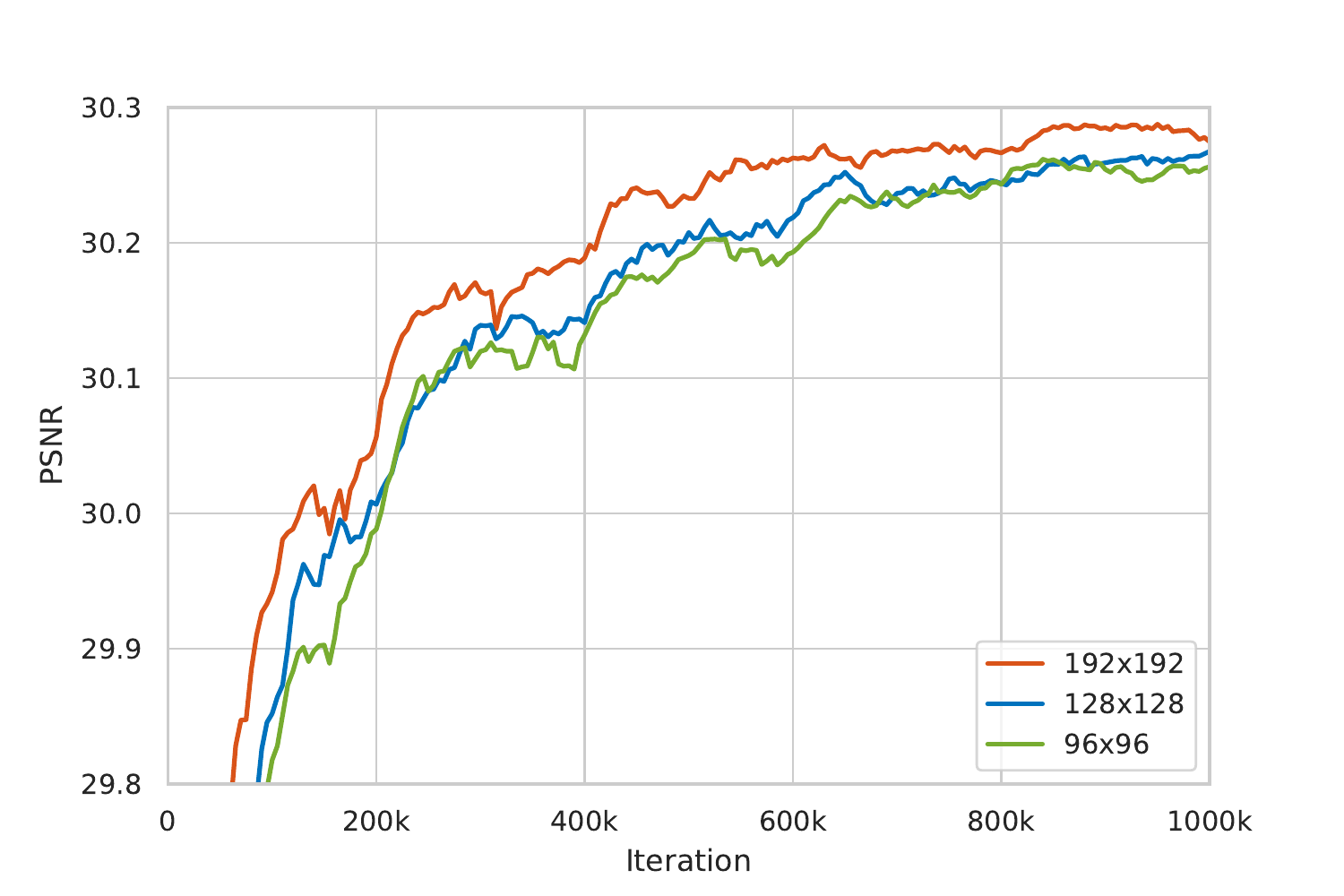}}}%
	\qquad
	\subfloat[23 RRDBs 
	]{{\label{fig:patch_size_RRDB}\includegraphics[width=.45\linewidth]{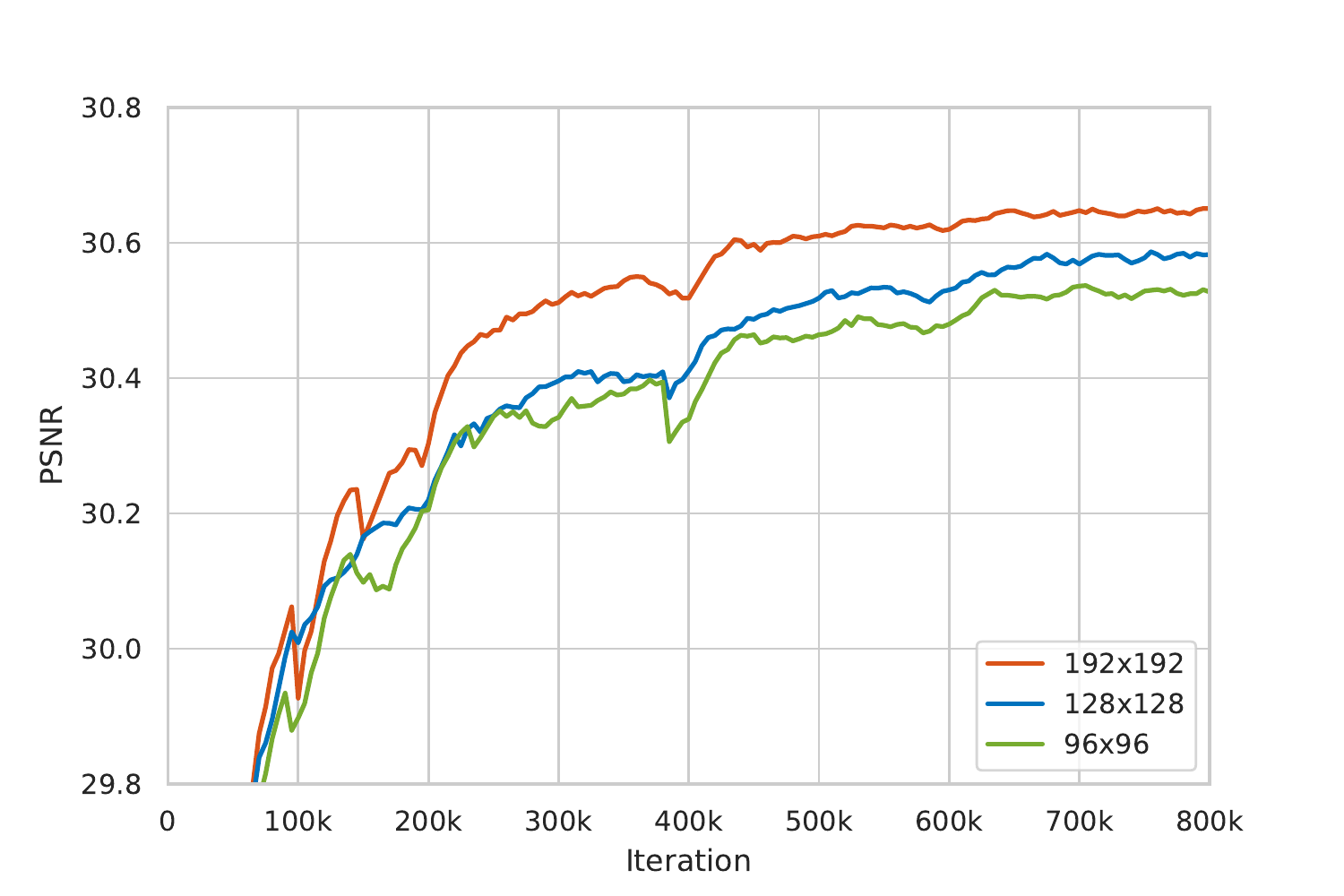}}}%
	\caption{The influence of training patch size (PSNR is evaluated on Set5 with RGB \mbox{channels}).}%
	\label{fig:patch_size}%
	\vspace{-0.5cm}
\end{figure}

We observe that training a deeper network benefits from a larger patch size, since an enlarged receptive field helps 
the network to capture more semantic information. 
We try training patch size $96\times96$, $128\times128$ and $192\times192$ on models with 16 RBs and 23 RRDBs (larger 
model capacity). The training curves (evaluated on Set5 with RGB channels) are shown in Fig.~\ref{fig:patch_size}. 

It is observed that both models benefit from larger training patch size.
Moreover, the deeper model achieves more improvement ($\sim$0.12dB) than the shallower one ($\sim$0.04dB) since
larger model capacity is capable of taking full advantage of larger training patch size.

However, larger training patch size costs more training time and consumes more computing resources. 
As a trade-off, we use $192\times192$ for PSNR-oriented methods and $128\times128$ for perceptual-driven methods.

\section{More qualitative comparison} \label{sec:qualitative_cmp}

\begin{figure}[htbp]
	\begin{center}
		\includegraphics[width=1\linewidth]{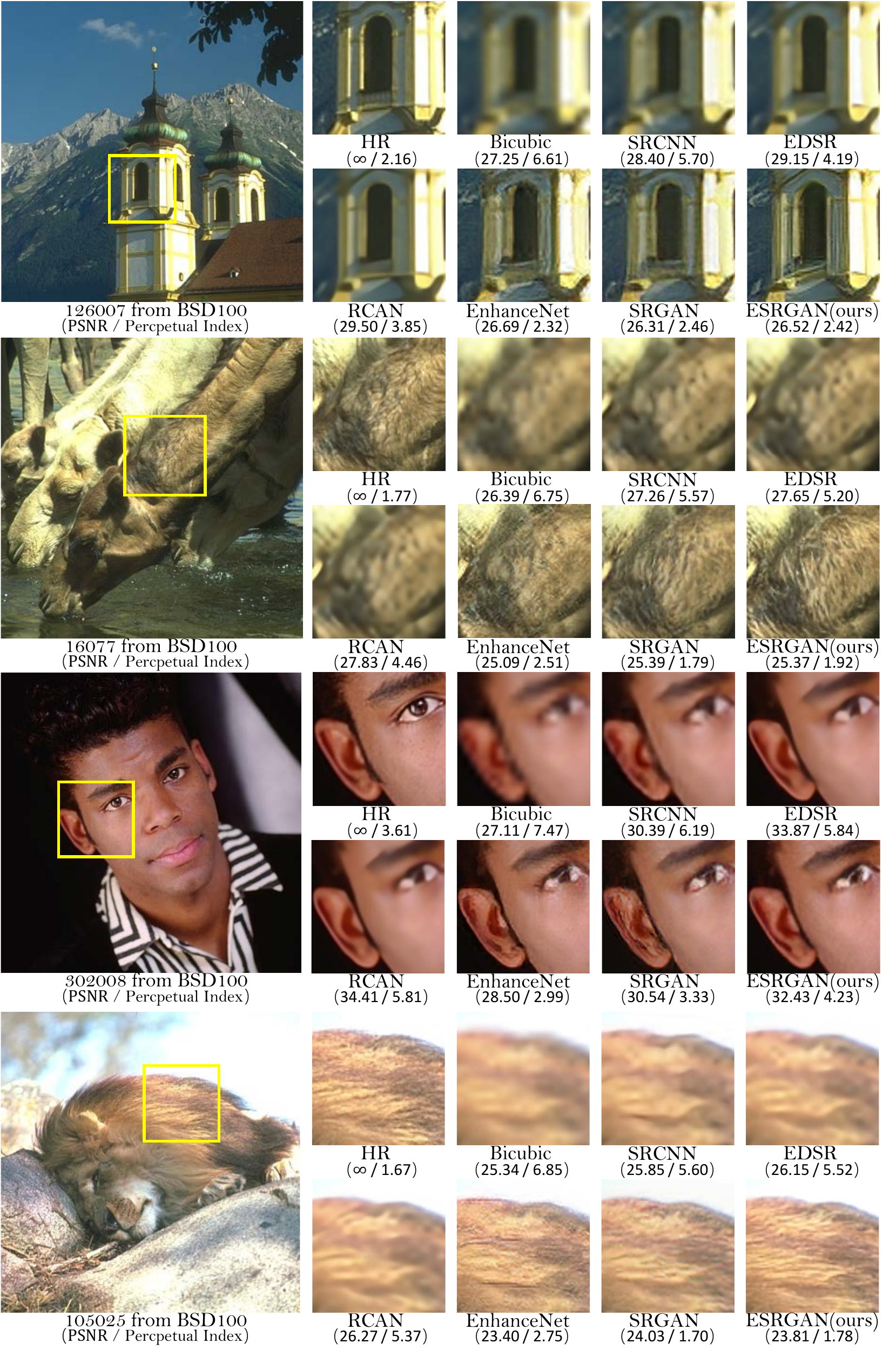}
	\end{center}
	\vspace{-0.4cm}
	\caption{More qualitative results. PSNR (evaluated on the Y channel) and the 
		perceptual index are also provided for reference.}
	\label{fig:intro_cmp}
\end{figure}

\end{document}